\documentclass[10pt,twocolumn,letterpaper]{article}

\usepackage{iccv}
\usepackage{times}
\usepackage{epsfig}
\usepackage{graphicx}
\usepackage{amsmath}
\usepackage{amssymb}

\usepackage{subcaption}
\usepackage[font=small]{caption}
\usepackage{algorithm, algpseudocode}
\usepackage{multirow}
\usepackage{bbm}
\usepackage{bm}

\makeatletter
\newcommand*\bigcdot{\mathpalette\bigcdot@{.5}}
\newcommand*\bigcdot@[2]{\mathbin{\vcenter{\hbox{\scalebox{#2}{$\m@th#1\bullet$}}}}}
\makeatother

\usepackage[pagebackref=true,breaklinks=true,letterpaper=true,colorlinks,bookmarks=false]{hyperref}

\iccvfinalcopy

\def\iccvPaperID{5232}

\ificcvfinal\fi

\begin{document}

\title{EgoPCA: A New Framework for Egocentric Hand-Object Interaction Understanding}

\author{
Yue Xu$^1$,~~
Yong-Lu Li$^{1,2}$\thanks{Corresponding author.},~~
Zhemin Huang$^1$,~~
Michael Xu Liu$^3$,~~
Cewu Lu$^1$,\\
Yu-Wing Tai$^4$,~~
Chi-Keung Tang$^2$
\\
{\small 
$^1$Shanghai Jiao Tong University~~
$^2$HKUST~~
$^3$New Hope Investment Group~~
$^4$Dartmouth College}
\\
{\tt\small \{silicxuyue, yonglu\_li, lucewu\}@sjtu.edu.cn} \\
{\tt\small zhemin.huang@outlook.com, Michaelliu@newhope.cn, yuwing@gmail.com, cktang@cs.ust.hk}
}


\twocolumn[{%
\renewcommand\twocolumn[1][]{#1}%
\maketitle
\begin{center}
    \centering
    \captionsetup{type=figure}
    \includegraphics[width=.71\textwidth]{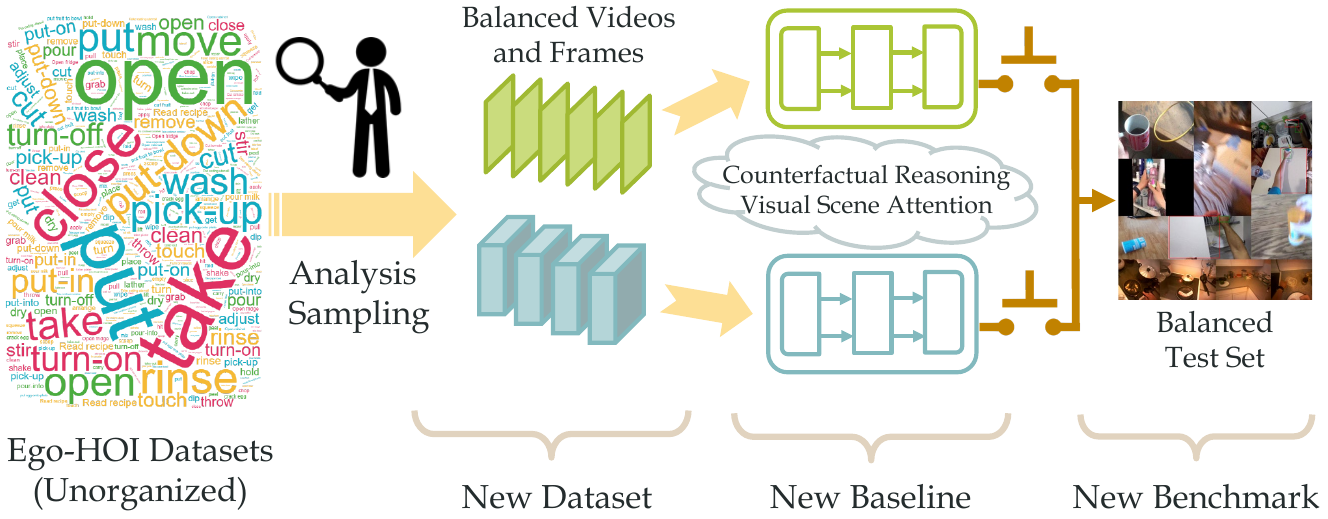}
    \captionof{figure}{A new framework for Ego-HOI learning, consisting of a new benchmark, a new baseline model, and a learning mechanism.}
    \label{fig:teaser}
\end{center}%
}]

{
  \renewcommand{\thefootnote}%
    {\fnsymbol{footnote}}
  \footnotetext[1]{Corresponding author.}
}
{
    \renewcommand{\thefootnote}{\fnsymbol{footnote}}
    \footnotetext[2]{This research is supported in part by the Research Grant Council of the Hong Kong SAR under grant no. 16201420.}
}

\ificcvfinal\thispagestyle{empty}\fi

\begin{abstract}
With the surge in attention to Egocentric Hand-Object Interaction (Ego-HOI), large-scale datasets such as Ego4D and EPIC-KITCHENS have been proposed.
However, most current research is built on resources derived from third-person video action recognition. This inherent domain gap between first- and third-person action videos, which have not been adequately addressed before, makes current Ego-HOI suboptimal. 
This paper rethinks and proposes a new framework as an infrastructure to advance Ego-HOI recognition by \textbf{P}robing, \textbf{C}uration and \textbf{A}daption (\textbf{EgoPCA}). We contribute comprehensive pre-train sets, balanced test sets and a new baseline, which are complete with a training-finetuning strategy. 
With our new framework, we not only achieve state-of-the-art performance on Ego-HOI benchmarks but also build several new and effective mechanisms and settings to advance further research. We believe our data and the findings will pave a new way for Ego-HOI understanding. 
\textbf{Code and data are available at \url{https://mvig-rhos.com/ego_pca}}.
\end{abstract}

\section{Introduction}
Understanding Egocentric Hand-Object Interaction (Ego-HOI) is a fundamental task for computer vision and embodied AI. 
To promote HOI learning, many egocentric video datasets~\cite{ego4d, epic, egtea, sthsth, charadesego} have been released, which contributed to recent advances in this direction.
Recently, deep learning based methods~\cite{tsn, slowfast, i3d}, especially Transformers and visual-language models~\cite{mvit, vivit, timesformer, actionclip} have achieved high performances on these benchmarks.

Though significant progress has been made, challenges remain. With few better choices available, current studies on Ego-HOI typically adopt existing tools and settings of third-person action recognition, despite the significant domain gap between egocentric and exocentric action~\cite{charadesego, ego-exo}.
Notably, third-person action depicts almost full human body and associated poses, while first-person action typically only engages hands; third-person videos are usually stable or readily stabilized, while first-person videos can exhibit different degrees of camera motion and shaking, which are possibly intended by the actor. 
Given the large domain gap, existing methods inherited from third-person vision are arguably unsuitable.
Moreover, it remains unclear whether the existing Ego-HOI datasets~\cite{ego4d,epic,egtea,sthelse,charadesego} can support the model pre-training for transferability on downstream tasks.
Thus, here, we address the important technical question in Ego-HOI:
\textbf{What are the effective model and training mechanisms for Ego-HOI learning?}

To understand the need for a new baseline and customized training for Ego-HOI learning, we analyze the existing paradigm and observe three main weaknesses:
\textbf{1)} Previous methods are mainly based on models pre-trained on Kinetics~\cite{kinetics400}. It has been widely discussed that third-person action datasets like Kinetics have a huge domain and the semantic gap with egocentric videos~\cite{charadesego,sigurdsson2018actor,yu2019see,choi2020unsupervised,ego-exo}. Thus, we need a new pre-train set specifically designed for Ego-HOI;
\textbf{2)} Previous ad-hoc models are designed for third-person video learning.
And these solutions are typically tailored to address one or a limited subset of Ego-HOI learning instead of a more general one-for-all model, \ie, one model for all Ego-HOI benchmarks;
\textbf{3)} In current schemes, finetuning one shared pre-trained model for all downstream tasks is inefficient, which also falls short of adapting to every downstream task or benchmark.
Therefore, a task-specific scheme is necessary so that we can efficiently learn a customized model for each downstream task. 

In light of these weaknesses, in this work, we propose a novel basic framework for Ego-HOI learning by Probing, Curation and Adaption (EgoPCA): we probe the properties of Ego-HOI videos, based on which we leverage data curation for balanced pre-train and test datasets, and finally adapt the model according to specific tasks. The details are as follows.

\textbf{1)} \textbf{New Pre-Train and Test Sets.}
We build a \textit{new comprehensive pre-train set} based on the videos from Ego-HOI datasets.
Although multiple datasets are available for training a universal Ego-HOI model, the noisy, highly long-tailed source datasets (\eg, EPIC-KITCHEN~\cite{epic} and EGTEA Gaze+~\cite{egtea}) can introduce imbalance to the pretrained models and thus adversely influence their generalization ability~\cite{entezari2023role}. The bulky ``head" data in the long-tailed distribution also result in unmanageable training cost given the current rapid growth of the model and data size. 
Hence we propose to seek a balanced pretrain data distribution for the training efficacy and efficiency.
In the scope of Ego-HOI, the data should be balanced not only on the semantics of samples but also on the other video properties such as camera motion or hand poses.
After conducting thorough studies on Ego-HOI video properties, we sample a small but balanced and informative subset from multiple datasets~\cite{ego4d,epic,egtea,sthelse}.
which can support better transfer learning for downstream tasks with domain and semantic gaps.
Alongside, \textit{a new balanced test set} is built that accounts for the long-tailed distribution of Ego-HOI videos and its HOI semantics for fair and unbiased evaluation of models, which is a widely adopted approach~\cite{openlongtail}.

\textbf{2)} \textbf{One-for-All Baseline Model}.
We propose a new baseline given the unique egocentric video properties, that consist of an efficient lite network and a representative heavy network, which can leverage both frames and videos in training.
Moreover, we observe that the camera motion associated with Ego-HOI videos often correlates to serial attention to the visual scene of interaction. So we propose {\em Serial Visual Scene Attention} (\textbf{SVSA}) prediction task to exploit such knowledge.
In particular, we incorporate \textit{counterfactual} reasoning in ego-videos, applying intervention on the ``hand'' causal node by replacing the hand patch with different hand states while keeping the scene/background. The model output should change after intervention.
With these constraints, our baseline achieves state-of-the-art (SOTA) on several benchmarks when pre-trained on our training set, and outperforms the SOTA significantly on our test set.

\textbf{3)} \textbf{All-for-One Customized Mechanism.}
Towards the best settings for each downstream task, we propose a new video sampling and selection algorithm based on the ego-video properties analysis. Given our one-for-all model, we apply our optimal training and tuning policies for each task. 
Subsequently, we further outperform the performance of our one-for-all model on several benchmarks.

Overall, to ``standardize'' Ego-HOI learning and integrate resources, our contributions are:
1) we revisit the Ego-HOI tasks and analyze the data from the perspectives of dataset construction and model design;
2) according to our analysis, instead of directly using the third-person video methods/tools, we propose a new framework (pre-train set, baseline, and test set) designed exclusively for Ego-HOI;
3) to pursue SOTA while minimizing training costs, we propose a customized approach for downstream tasks.

\section{Related Work}

\subsection{HOI Understanding}

Different from the third-person and general HOI learning~\cite{pastanet,tincvpr} that studies the interactions between the whole body and object, Ego-HOI only focuses on the hand-object interactions in the egocentric view.
Recently, various Ego-HOI datasets have been proposed~\cite{ego4d, epic, egtea, sthsth, charadesego}. 
EPIC-KITCHENS~\cite{epic} is one of the first large-scale Ego-HOI datasets with over 80 K instances, more general actions, and objects, where hand and object positions are available. 
Similarly, EGTEA Gaze+~\cite{egtea} contains HOI annotation and provides auxiliary gaze data. 
The success of deep learning has promoted the development of Ego-HOI recognition models, including 2D ConvNets~\cite{tsn}, multi-stream networks~\cite{slowfast}, and 3D ConvNets~\cite{i3d}. Transformer-based networks and visual-language models have played important roles in HOI learning~\cite{mvit, movinet, actionclip}, which can significantly boost performance.

\subsection{Video Action Recognition}
Video action recognition is a foundational task.
In terms of \textbf{backbones}, previous methods can roughly fall into three groups. 
\textit{Two-stream}~\cite{two-stream, two-stream1, two-stream2},
\textit{3D CNN}~\cite{i3d, r21d, s3d, slowfast},
and 
\textit{Transformer} methods~\cite{timesformer, vivit, mvit, memvit, swin}.
In terms of large-scale \textbf{pre-training}, it has already been a standard procedure for Ego-HOI models.
In the early ages, researchers leverage the pre-training on large-scale image datasets, \eg, ImageNet~\cite{imagenet1, imagenet2} and MS-COCO~\cite{mscoco}. 
Some methods follow the image pre-training and extract features from video frames, while such methods can not exploit the temporal information and require aggregation of such information.
Given large-scale video datasets, methods~\cite{Purwanto_2019_ICCV,Girdhar_2019_CVPR, gabeur2020multi,li2020hero,li2021bridging} pre-train models on Kinetics~\cite{kinetics400, i3d} or HowTo100M~\cite{howto100m} to utilize the transferability and enhance recognition. 
Besides, CLIP~\cite{clip} is a milestone in adopting contrastive learning with large-scale image-text pairs, demonstrating outstanding zero-shot performance.
In terms of \textbf{transfer learning}, previous works typically finetune the pre-trained backbone paired with a new classifier to adapt to downstream tasks. 
ActionCLIP~\cite{actionclip} end-to-end finetunes on target datasets and shows that finetuning is critical to both language and image encoders. 
Ego-Exo~\cite{ego-exo} uses Kinetics pre-trained backbone and finetunes it on the target egocentric dataset.

\section{Ego-HOI Videos}

Egocentric videos have various properties from exocentric videos as they are often characterized by more camera motions, higher blurriness, \etc.
So we explore the key properties inherent in Ego-HOI videos to guide the framework design (Section~\ref{sec:prop_ana}) and  
propose our sampling strategy for balanced data.
We introduce the ego-property similarity and selection method (Section~\ref{sec:semantic}) and
the construction of our pre-train set {\bf One4All-P} (Section~\ref{sec:one4all-pretrain}), which is comprehensive, generalizable and transferable.
Finally, we construct our balanced test set {\bf One4All-T} (Section~\ref{sec:one4all-test}).

\subsection{Ego-HOI Video Properties}
\label{sec:prop_ana}

We first present how to quantitatively measure the video properties to visualize and derive our sampling strategy. 
The comprehensive analysis will be presented on five datasets: EPIC-KITCHENS-100~\cite{epic}, EGTEA Gaze+~\cite{egtea}, Ego4D-AR~\footnote{
Ego4D-AR (\textbf{A}ction \textbf{R}ecognition) is constructed based on the hand-object interaction split of Ego4D~\cite{ego4d}. We assign the action labels from anticipation tasks to the video clips for the HOI task to build an action recognition benchmark. Please refer to the supplementary for more details.
}, Something-Else~\cite{sthelse} and our One4All-P.

\begin{figure}[t]
    \centering
    \includegraphics[width=0.99\linewidth]{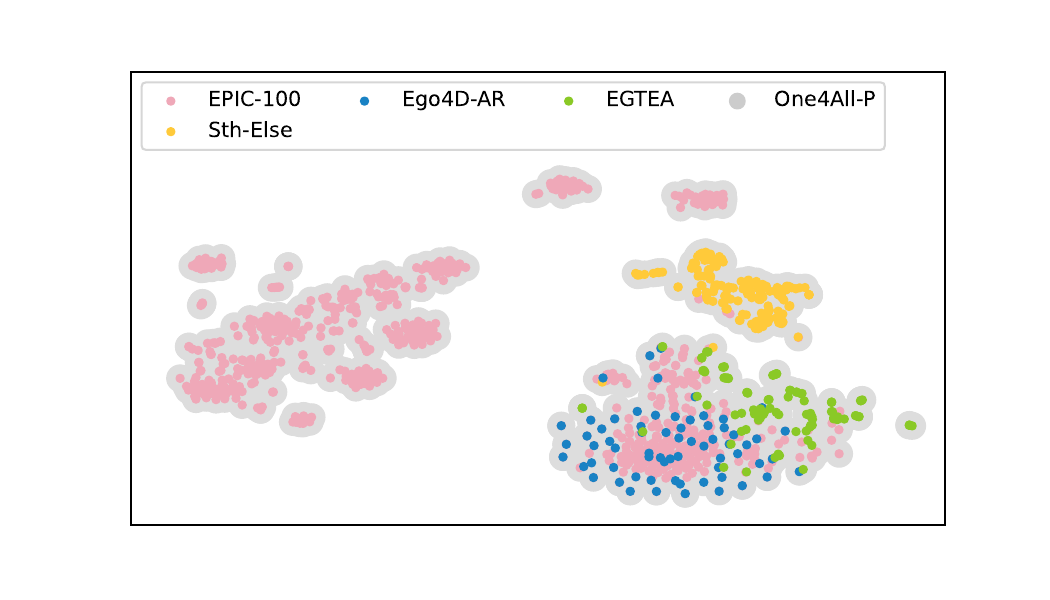}
    \caption{Semantic distribution of actions of Ego-HOI \textbf{train} sets. We use BERT~\cite{devlin2018bert} embeddings to visualize the classes.}
    \label{fig:semantic_action}
\end{figure}

{\bf Ego-HOI Semantics.}
The action label of a video clip is one of the most important properties of egocentric videos.
Considering the ego-property similarity in the labels among different datasets, we represent the HOI semantics of a video clip as the label word vectors extracted by the pre-trained BERT~\cite{devlin2018bert}. Thus, videos with similar HOIs are in close proximity in the BERT latent space.
Figure~\ref{fig:semantic_action} depicts the t-SNE visualization of the class semantics of several datasets, which shows that our One4All-P spans the largest area and that the class embeddings of Something-Else are differently located from the rest of the datasets.

\begin{figure}[t]
    \centering
    \includegraphics[width=\linewidth]{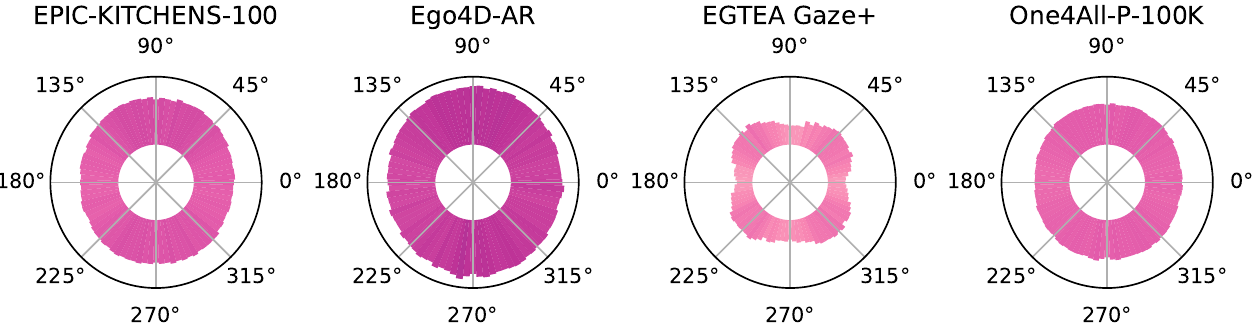}
    \caption{Camera motion polar histogram of Ego-HOI \textbf{train} sets. The length and angle of the bars: the motion magnitude and angle.}
    \label{fig:camera_motion}
\end{figure}

{\bf Camera Motion.}
Different from third-person videos typically shot by stable cameras, egocentric videos are captured by wearable cameras, so they exhibit a wider variety of viewpoints, view angles, and shaking movements. Such camera motions highly correlate to the human's intention in the HOI task which helps video understanding, \eg, which object to interact with in the next step.
We use dense optical flows between frames to quantify per-pixel camera motion. We compute the polar histogram of shift vectors by angles and take the angle and length of the largest bin to represent the camera motion of the frame. 
The camera motion of each video is represented as the polar histogram of motion vectors of the frames. 
Figure~\ref{fig:camera_motion} shows the polar histogram of camera motion of datasets, and EPIC-100 and Ego4D-AR exhibit larger motion than EGTEA and One4All-P.

\begin{figure}[t]
    \centering
    \includegraphics[width=0.99\linewidth]{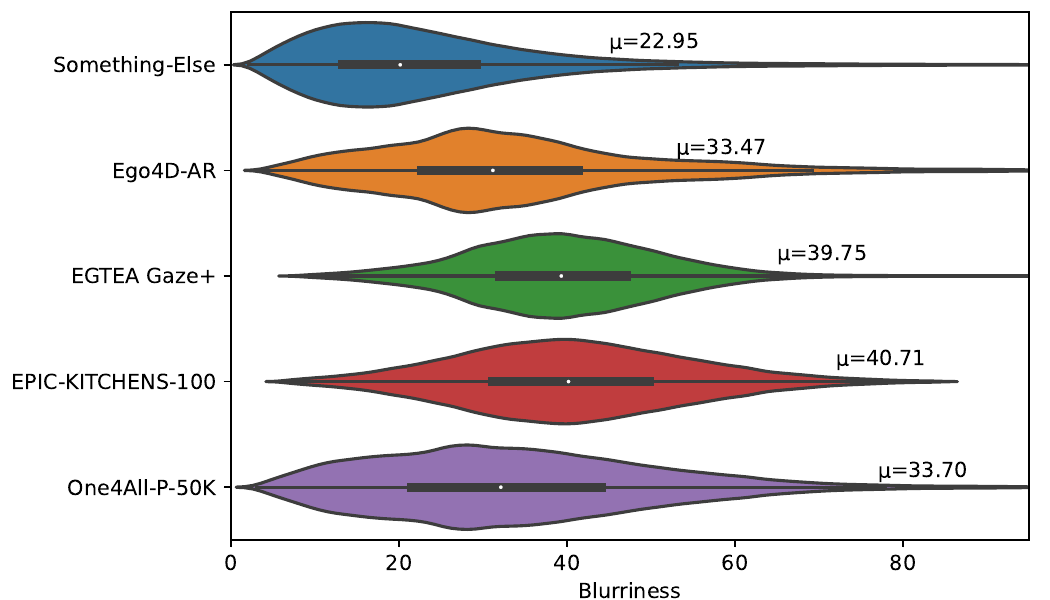} 
    \caption{Blurriness (train sets). $\mu$: average blurriness value.}
    \label{fig:blur}
\end{figure}

{\bf Blurriness.}
Egocentric videos can be blurry due to fast camera motion, either intentional or occasional. 
We measure the blurriness by the variance of Laplacian of the frames since a \textit{blurry} frame has a \textit{smaller} variance of Laplacian. Then each video is represented as the mean and variance of blurriness of multiple frames.
Figure~\ref{fig:blur} shows the distribution of blurriness. 
Something-Else has the lowest blurriness score as it was captured with less camera movement. In comparison, our One4All-P is more balanced in blurriness and covers the blurriness ranges of other datasets.

\begin{figure*}[t]
    \begin{minipage}[b]{0.32\linewidth}
        \centering
        \includegraphics[width=\textwidth]{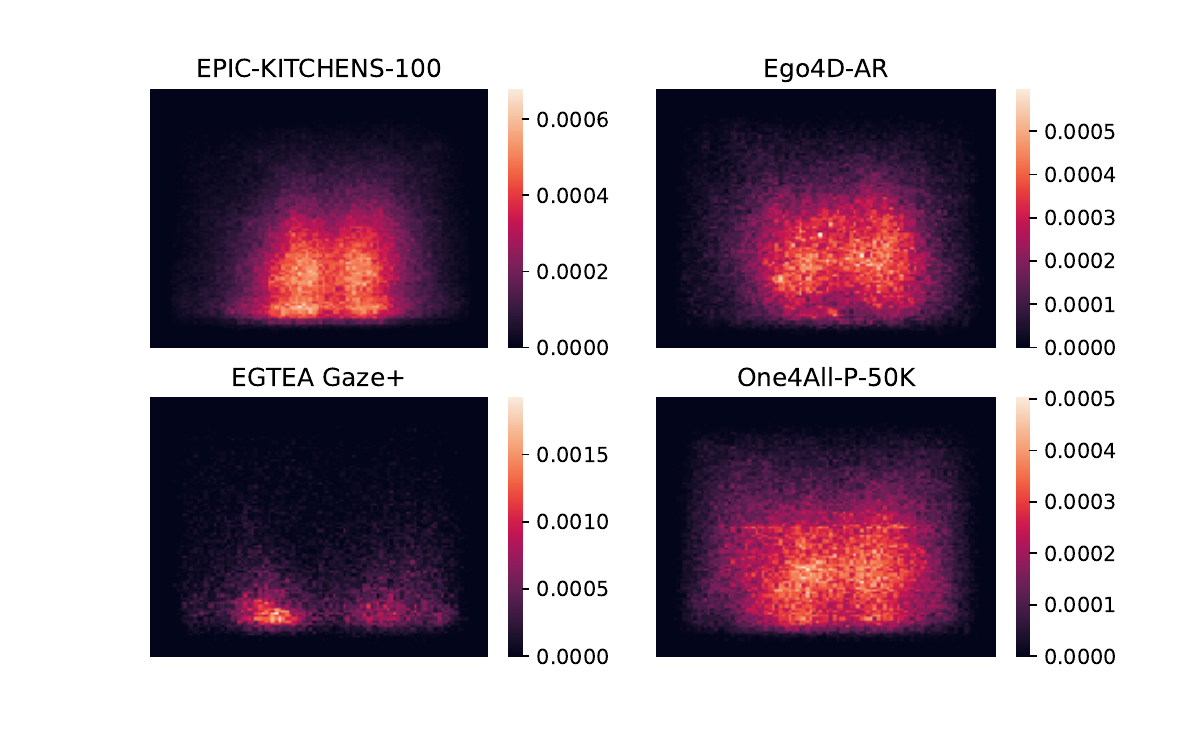} 
        \caption{Hand location heatmaps of Ego-HOI datasets (\textbf{train} set).}
        \label{fig:hand_box}
    \end{minipage}
    \hfill
    \begin{minipage}[b]{0.32\linewidth}
        \centering
        \includegraphics[width=\textwidth]{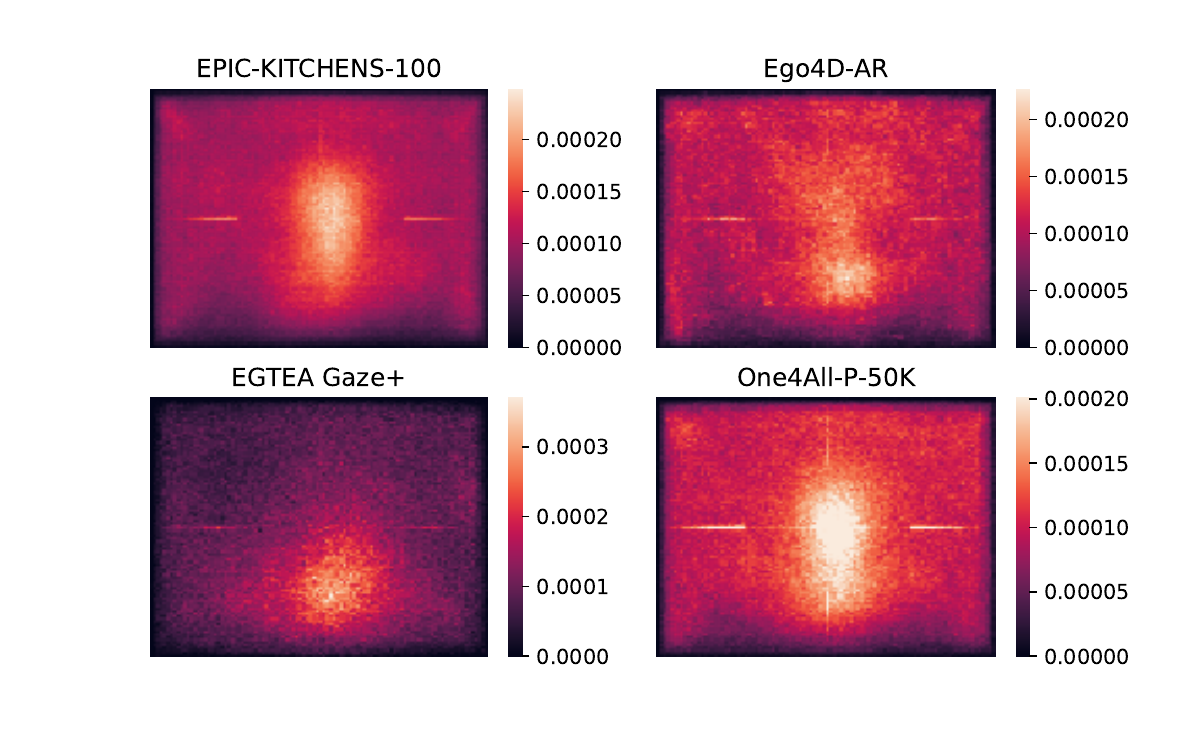} 
        \caption{Object location heatmaps of Ego-HOI datasets (\textbf{train} set).}
        \label{fig:obj_box}
    \end{minipage}
    \hfill
    \begin{minipage}[b]{0.32\linewidth}
        \centering
        \includegraphics[width=\textwidth]{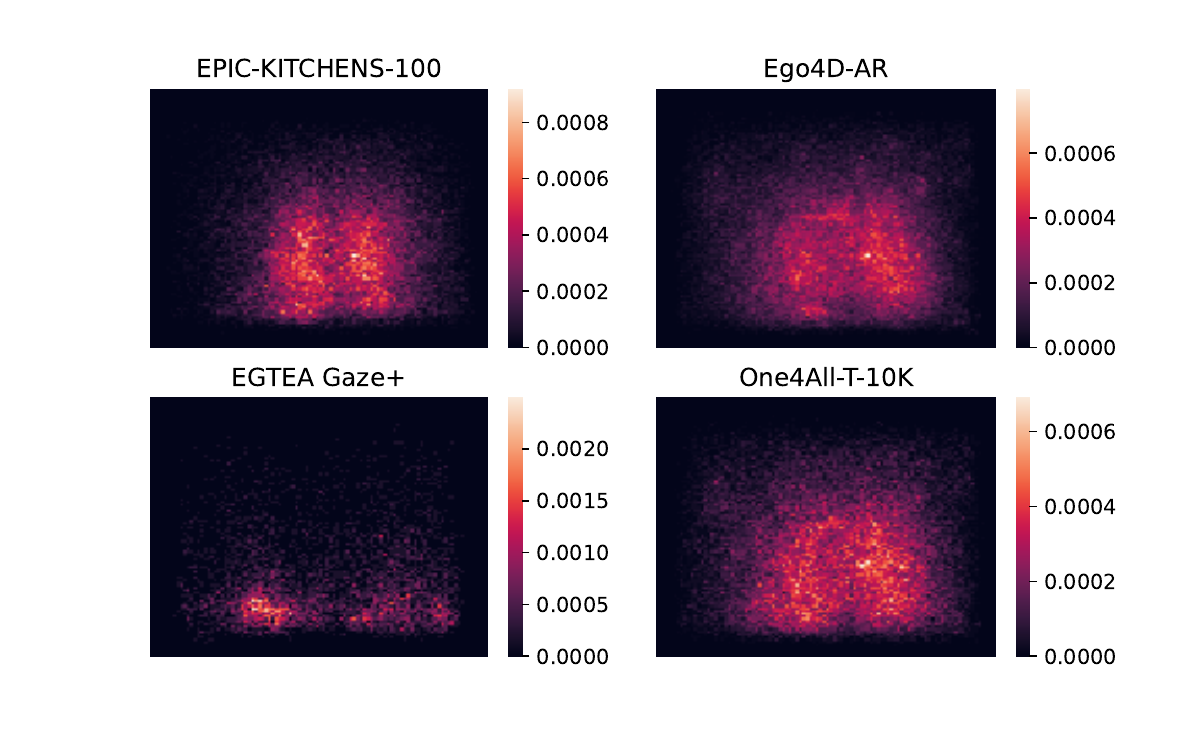} 
        \caption{Comparison of hand location of \textbf{test} set of Ego-HOI datasets.}
        \label{fig:val_hand_location}
        
    \end{minipage}
\end{figure*}

{\bf Hand/Object Location.}
The location distribution of hands or objects varies among different datasets. 
We extract the hands and objects' location with existing detection toolboxes (MMPose~\cite{mmpose} for hand and Detic~\cite{detic} for object), and each video is represented as discrete heatmaps of hands and objects.
We show the heatmaps of hand and object locations (Figure~\ref{fig:hand_box}, \ref{fig:obj_box}). The hands and objects are primarily located at the bottom of the frames in EGTEA Gaze+, while they are rather arbitrary in EPIC-100 and One4All-P.

\begin{figure}[t]
    \centering
    \includegraphics[width=\linewidth]{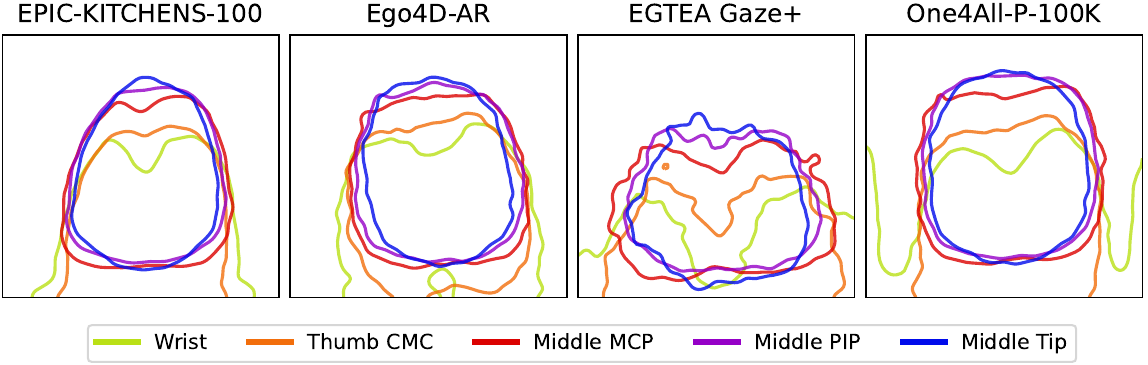} 
    \caption{Hand pose. We show the high-density contours of the heatmaps of different hand keypoints on different \textbf{train} sets.
    }
    \label{fig:hand_pose}
\end{figure}

\begin{figure}[t]
    \centering
    \includegraphics[width=\linewidth]{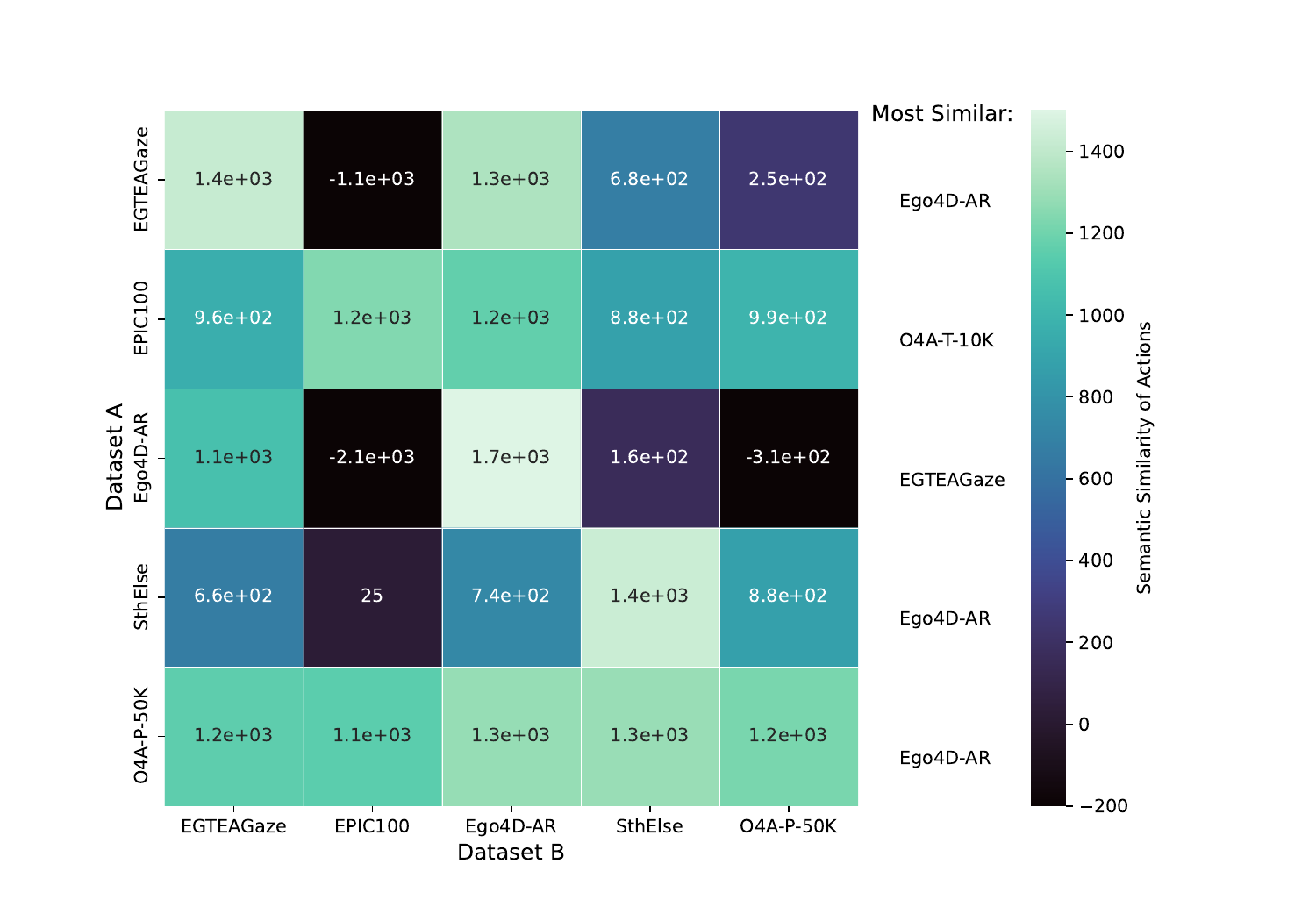}
    \caption{The unified ego-property similarity between datasets.}
    \label{fig:semantic_dis}
\end{figure}

{\bf Hand Pose.}
To more accurately capture the detail of human hands, we measure the hand pose in videos with an off-the-shelf pose detector (MMPose~\cite{mmpose}). Then, each video clip is represented as the 21 keypoints (a 42-dim vector).
Figure~\ref{fig:hand_pose}
visualizes the distribution of hand poses, where we generate heatmaps of 5 main keypoints (from wrist to middle fingertip) and draw the contours of their high-density area. The figure shows that the hands are usually placed vertically and fingers pointing upwards in Ego-HOI datasets, which is omitted in hand box representations. Moreover, the hands in EPIC-100 are closer to the center than those of EGTEA Gaze+. Compared to EPIC-100 and Ego4D, the hand pose of One4All-P is more diversified.

We also present comparisons between valid/test sets of Ego-HOI datasets.
The hand location heatmaps are shown in Figure~\ref{fig:val_hand_location}. For the rest, please refer to the supplementary. Our One4All-Val also shows balancedness on the proposed video properties over other datasets.

\subsection{Ego-Property Similarity}
\label{sec:semantic}
To measure the property similarity between datasets, we propose the \textbf{ego-property similarity}. 
For the similarity between sets $A$ and $B$, we use Kernel Density Estimation (KDE) to estimate the distribution of $A$ as $\widetilde{P_A}$. Then the similarity is measured as the likelihood of the set $B$ on $P_A$:
\begin{equation}
    \text{Sim}(A, B) = \widetilde{P_A}(B) = \prod_{x \in B} \widetilde{P_A}(x),
\end{equation}
where the representation $x$ of a sample is one of the aforementioned quantitative properties, \eg semantic BERT vectors, hand pose keypoints.
\begin{algorithm}[ht]
    \caption{Video Selection}
    \label{alg:sampling-algorithom}
    \begin{algorithmic}[1]
        \Require Source data $S$ and extra data $E$, KDE update frequency $k$, target instance number $m$, temperature $\tau$
        \Ensure Selected instances $T=\{t_1, \cdots, t_m\}$
        \State $T\gets \{\}$
        \Repeat
            \State Train KDE model $P_S$ with source data $S$,
            \State Compute log-likelihood $q_i=\log P_S\left(e_i\right), \forall e_i \in E$,
            \State Compute sampling probability $p_i$ with Equation~\ref{eq:sem-sampling}
            \State Draw $k$ instances $E_k$ from $E$ with distribution $p_i$
            \State $S \gets S\bigcup E_k $, $T \gets T\bigcup E_k $
            \State $E \gets E \backslash E_k $
        \Until{$|S|$ reach $m$}
    \end{algorithmic}
\end{algorithm}
In KDE, we assume the Gaussian kernels have diagonal covariance, and the bandwidth of each dimension is selected with Silverman's estimator~\cite{silverman2018density}.
Exceptionally, for the blurriness, we regard the mean values of blurriness of video frames as the representation $x$ and the standard deviation as the bandwidth.
As an example, we compare the ego-property similarity between datasets in Figure~\ref{fig:semantic_dis} and list the most similar dataset for each dataset.

\begin{table*}[t]
\centering
    \resizebox{\textwidth}{!}{
          \begin{tabular}{l|c|cccccc|c}
          \hline
            & Random & Action Semantic & Camera Motion & Blurriness & Hand Location & Object Location & Hand Pose & Unified (best weight)\\ 
          \hline
          $+5\%$  & 69.7 (+0.3) & 70.0 (+0.5) & 70.5 (+1.0) & 70.0 (+0.5) & 70.5 (+1.0) & \underline{70.5 (+1.1)} & 70.3 (+0.9)  & \textbf{70.6 (+1.2)} \\
          $+10\%$ & 69.7 (+0.2) & 70.1 (+0.6) & 70.2 (+0.8) & 70.2 (+0.7) & \textbf{71.0 (+1.5)} & 69.7 (+0.3) & 70.0 (+0.6)  & \underline{70.6 (+1.2)} \\
          $+20\%$ & 69.6 (+0.2) & 70.2 (+0.8) & 70.3 (+0.8) & 70.0 (+0.5) & \underline{70.9 (+1.5)} & 70.3 (+0.8) & 70.2 (+0.7)  & \textbf{71.2 (+1.8)} \\
          \hline
          \end{tabular}
      } 
      \captionof{table}{Performance after adding data to EGTEA Gaze+~\cite{egtea} split 3 according to various criteria. The baseline (w/o additional data) accuracy is \textbf{69.4}\%. We find that the \textbf{camera motion, hand location/pose, and object location} are more important among all the factors. }
      \label{tab:semantic-sampling}
\end{table*}

{\bf Video Selection}.
Based on the ego-property similarity, we propose an \textbf{ego-property similarity-based selection algorithm} to sample extra data to enrich the original video set towards \textit{balancedness} or \textit{higher performance}. 
We estimate the KDE distribution $\widetilde{P_S}$ of source dataset $S$ and select a subset $T$ from extra dataset $E$ based on the likelihood $\widetilde{P_S}(e_i)$ of each sample $e_i$. 
If the aim is performance, we maximize the \textit{ego-property similarity} between $S$ and $T$, so the sampling probability is the normalized likelihood $p_i\propto \widetilde{P_S}(e_i)$. 
And if the aim is balancedness, we maximize the \textit{distance} between $S$ and $T$ for data diversity, so we incorporat reversed probability $p_i\propto \widetilde{P_S}(e_i)^{-1}$. Since the z-score normalization of probability is equal to the softmax of log-likelihood, the sampling probability is formulated:
\begin{equation}
p_i = \left\{
\begin{aligned}
& \text{softmax}(\frac1{\tau} \log \widetilde{P_S}(e_i)),  \quad\text{(performance)} \\
& \text{softmax}(- \frac1{\tau} \log \widetilde{P_S}(e_i)), \quad\text{(balancedness)} \\
\end{aligned}
\right.
\label{eq:sem-sampling}
\end{equation}
where temperature $\tau$ modulates the sampling strength. 
We gradually select samples and update the KDE distribution per $k$ instances. The input is our video property representations.
The complete algorithm is shown in Algorithm~\ref{alg:sampling-algorithom}.

We conduct an ablation study in Table~\ref{tab:semantic-sampling}. We add auxiliary videos from 3 other datasets (\cite{epic, sthelse, ego4d}) to EGTEA Gaze+ to enhance the performance according to different video properties. Results indicate that camera motion, hand location and pose, and object location are better criteria for video selection compared to semantics, blurriness, \etc.

\begin{figure}[t]
\centering

\begin{minipage}[b]{0.6\linewidth}
    \resizebox{\linewidth}{!}{
        \begin{tabular}{l|c|c}
        \hline
        Dataset & Sample & Class \\
        \hline
        EGTEA Gaze+~\cite{egtea}        & 8,300   & 106  \\
        Something-Else~\cite{sthelse}   & 157,389 & 174  \\
        EPIC-KITCHENS-100~\cite{epic}   & 67,217  & 97   \\
        Ego4D-AR~\cite{ego4d}           & 22,081  & 66   \\
        Kinetics-400~\cite{kinetics400} & 306,245 & 400  \\
        \hline
        One4All-P-20K     &  20,000  & 394 \\
        One4All-P-30K            &  30,000   & 394 \\
        One4All-P-50K            &  50,000   & 394 \\
        \hline
        \hline
        EGTEA Gaze+~\cite{egtea} (test split 3) & 2,021  & 106 \\
        Something-Else~\cite{sthelse} (val)     & 22,660 & 174 \\
        EPIC-KITCHENS-100~\cite{epic} (val)     & 9,668  & 97  \\
        Ego4D-AR~\cite{ego4d} (val)             & 14,530 & 58  \\
        \hline
        One4All-T-3K  &  3,000    & 204 \\
        One4All-T-5K         &  5,000    & 204 \\
        One4All-T-10K        &  10,000   & 204 \\
        \hline
        \end{tabular}
        } 
    \subcaption{Datasets}

\end{minipage}
\hfill
\begin{minipage}[b]{0.37\linewidth}
    \centering
    \includegraphics[width=\linewidth]{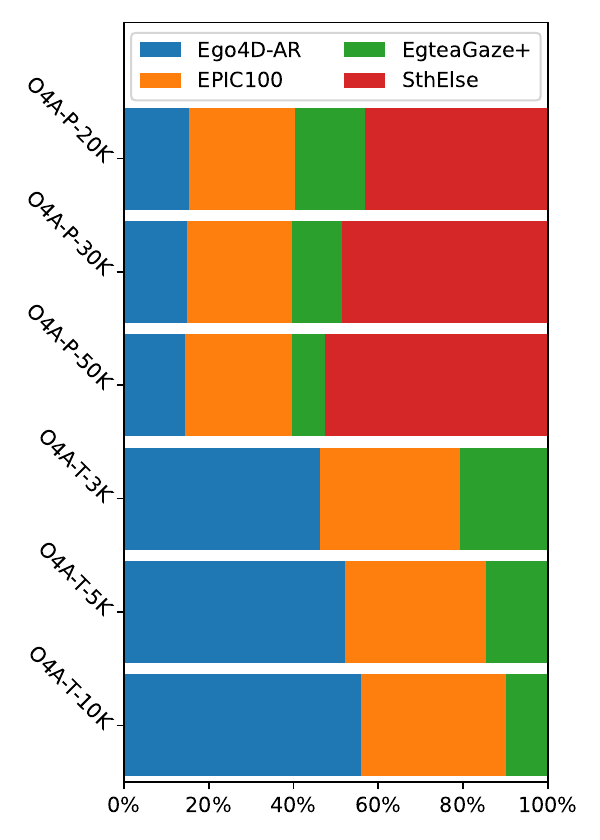} 
    \subcaption{Data Sources}
\end{minipage}

\captionof{table}{Previous datasets and our pre-train/test sets. (a) The upper block indicates the (pre)train sets and the lower block shows the validation/test sets. (b) The components in our datasets.}
\label{tab:pretrainset}
\end{figure}

{\bf Unified Ego-Property Based Sampling}.
Then we propose a unified sampling criterion with ego properties. We compute the sampling probability with Eq.~\ref{eq:sem-sampling} for each video property and take their \textit{weighted sum} as unified sampling probability. 
The weight is obtained in proportion to the \textit{significance of each property}. 
An example is given in Table~\ref{tab:semantic-sampling}.
The ablation shows the superiority of the unified criterion.

\subsection{Constructing A Comprehensive Pre-train Set}
\label{sec:one4all-pretrain}
Considering the generalization and transfer ability of the pre-trained model, we intend for a more \textbf{comprehensive} pre-train set, which is \textbf{balanced} and \textbf{diverse} not only on \textit{labels} but also on the proposed \textit{properties}.
With the ego-property similarity and selection algorithm, we progressively select samples to enhance the dataset's balancedness on multiple properties while ensuring high sample diversity. We construct our pre-train set with EPIC-KITCHENS-100, EGTEA Gaze+, Something-Else, and Ego4D-AR.
Note that we only use Something-Else for training since it also contains third-person videos (roughly 10\%), in order to exploit its high diversity in hand and object.
We first merge the action labels of the datasets and merge the semantically identical classes. Then we randomly sample 30 instances for each class for a class-balanced base dataset. 
Then, the Algorithm~\ref{alg:sampling-algorithom} is applied to complement the dataset while keeping the balancedness.
Thus, we propose our pre-train datasets \textbf{One4All-P-20K}, \textbf{One4All-P-30K}, and \textbf{One4All-P-50K}, with 20 K, 30 K, and 50 K video clips respectively.
Table~\ref{tab:pretrainset} shows their details.
Here, we aim at studying how to build a high-quality while minimal pre-train set to improve efficiency while pursuing maximum performance. Adding more data may indeed improve performance while sometimes may degrade the scores, as the more severe longtailed distribution, the less diversity, background bias, \etc.

\begin{figure*}[t]
    \centering
    \includegraphics[width=0.99\linewidth]{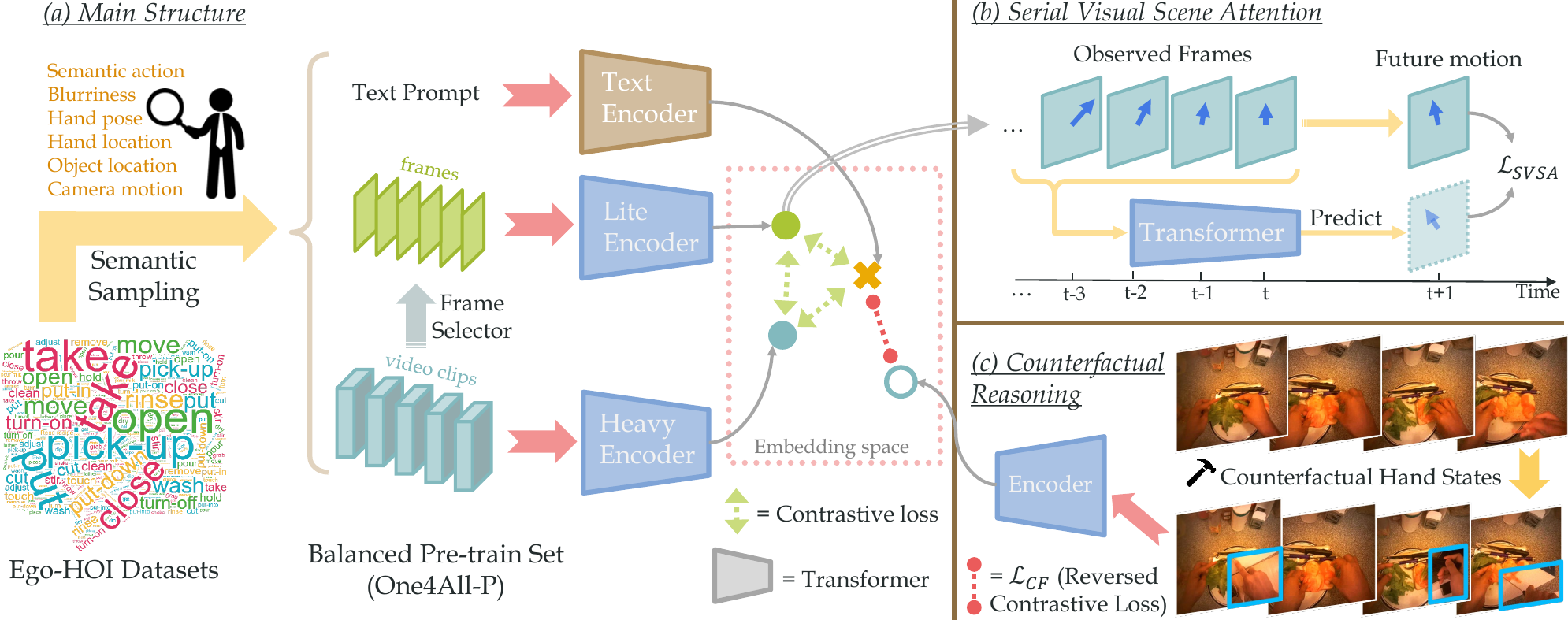} 
    \caption{Baseline model overview. (a) It consists of a lite and a heavy network. The embeddings are aligned with text features by contrastive learning; (b, c) Given the unique properties, we propose SVSA and counterfactual reasoning to promote Ego-HOI learning.}
    \label{fig:model}
\end{figure*}

\subsection{Constructing A More Balanced Test Set}
\label{sec:one4all-test}

The widely adopted Ego-HOI benchmarks like EPIC-KITCHENS and EGTEA-Gaze+ are either limited in scale or possess severely long-tailed test sets, resulting in a skewed evaluation.
Thus, a more balanced test set is required by the Ego-HOI community for a \textbf{fair} and \textbf{balanced} evaluation, which is balanced from multiple aspects like interaction semantics, hand/object locations, \etc
We use the same video selection approach (Algorithm~\ref{alg:sampling-algorithom}) to extract our new test sets \textbf{One4All-T-5K}, \textbf{One4All-T-10K}, and \textbf{One4All-T-20K}. Table~\ref{tab:pretrainset} tabulates the details of the test set.

\section{Methodology}

We propose our paradigm based on the analysis.
Existing methods typically adopt approaches of third-person action. 
Considering the gap between egocentric and exocentric HOI, we propose a baseline and pre-train it on our pre-train dataset for a \textbf{one-for-all} model (Section~\ref{sec:method:one4all}).
Then the pre-trained model can be finetuned to a stronger task-specific model with our customization (Section~\ref{sec:method:all4one}).

\subsection{One-for-All (One4All) Baseline Model}
\label{sec:method:one4all}

Ego-HOI data has unique properties making it unsuitable to use third-person models and pre-train directly. Moreover, these properties should be utilized rather than ignored in Ego-HOI models.
So we propose a new baseline model for Ego-HOI learning. As shown in Figure~\ref{fig:model}, our model resembles CLIP~\cite{clip} and consists of three encoders: lite, heavy, and text networks. The lite network captures frame-level features while the heavy network learns spatio-temporal features. These two streams are aligned with the text feature. As the instances in a batch may belong to the same class, we incorporate a KL contrastive loss following~\cite{actionclip} different from CLIP~\cite{clip}: in each $B$-sized batch, the output visual features $\mathbf{F} = \bm{f_i}|_{i=1}^B$ are aligned with the text feature $\mathbf{T} = \bm{t_i}|_{i=1}^B$ of label prompts by the loss:
\begin{equation}
\begin{aligned}
    \mathcal{L}_{kl}\left(\mathbf{F}, \mathbf{T}, \bm{y}\right) = 
    & \frac1{B}\sum_{i=1}^B KL [\mathit{Softmax}(\frac{ \bm{S}_{i \bigcdot} }{\tau})  \|  \bm{Q}_{i \bigcdot} ] + \\
    & \frac1{B}\sum_{j=1}^B KL [\mathit{Softmax}(\frac{ \bm{S}_{\bigcdot j} }{\tau})  \|  \bm{Q}_{\bigcdot j} ] ,
\label{eq:kl_loss}
\end{aligned}
\end{equation}
where $\bm{y}$ is the class label
and $\tau$ is the softmax temperature.
$\bm{S}_{ij} = \cos\langle \bm{f_i},~\bm{t_j} \rangle$ is the cosine similarity matrix between visual and text features. $\mathbf{Q}$ is the ground truth matrix and $\bm{Q}_{ij}$ is 1 only if $i^{th}$ and $j^{th}$ instance is in the same class. 
The KL contrastive loss draws closer to the visual and text features that have the same semantics.

\begin{table*}[t]
\centering
\resizebox{0.95\textwidth}{!}{
    \begin{tabular}{ll|c|ccc|ccc}
    \hline
    &Model               & Pre-train Set  &Ego4D-AR&EPIC-100& EGTEA  & One4All-T-3k &One4All-T-5K& One4All-T-10k\\ 
    \hline
    1& Chance                       & /               &  1.6  &   1.0  &  1.1  & 0.2 & 0.2  &  0.2  \\
    2& CLIP~\cite{clip}            & CLIP 400M     &  4.0  & 10.4 & 17.2 &  2.3 &  1.8 &  1.5 \\
    3& CLIP~\cite{clip}            & Kinetics-400  &  3.0  &  7.2 & 23.7 &  3.5 &  2.6 &  2.2 \\
    4& ActionCLIP~\cite{actionclip}& Kinetics-400  &  3.0  &  8.8 & 18.5 &  2.9 &  2.2 &  1.8 \\
    5& EgoVLP~\cite{egovlp}        & EgoClip       &  2.1  &  5.5 & 12.4 &  1.6 &  1.2 &  1.0 \\
    6& CLIP~\cite{clip}            & One4All-P-50K &  6.9  & 21.4 & 35.7 & 15.7 & 14.6 & 13.7 \\
    7& ActionCLIP~\cite{actionclip}& One4All-P-50K &  5.8  & 33.8 & 44.2 & 21.0 & 20.1 & 19.3 \\
    \hline
    8& Ours (Full)                 & Random-50K    &  6.9  & 34.6 & 50.5 & 22.7 & 20.7 & 20.2 \\
    9& Ours (Lite)                & One4All-P-50K &  5.6  & 40.5 & 48.9 & 23.2 & 22.2  & 21.4 \\
    10& Ours (Full)                & One4All-P-20K &  6.6  & 35.1 & 50.8 & 22.5 & 20.8 & 19.5 \\
    11& Ours (Full)                & One4All-P-30K &  6.0  & 35.7 & 52.5 & 22.7 & 20.9 & 19.7 \\
    12& Ours (Full)                & One4All-P-50K &  \textbf{7.2}  & \textbf{41.8} & \textbf{52.9} & \textbf{25.1} & \textbf{23.8} & \textbf{23.3} \\
    \hline
    \end{tabular}
} 
\captionof{table}{Performance of one-for-all model on benchmarks. Our lite model adopts the CLIP pre-trained model, and the heavy model uses a Kinetics-400 pre-trained MViT backbone. \textit{Full} means the simple late fusion of the lite and heavy model logits. Top-1 accuracy is reported.
} 
\label{tab:one4all}
\end{table*}

Specifically, the model is trained in multiple steps. First, the frame-level lite network is pre-trained with frame-text pairs in Ego-HOI data. Then we freeze the frame encoder and pre-train ATP module~\cite{atp} with video-text pairs. ATP is a keyframe selector that automatically selects the most informative frame given a batch of features. We sample $N$ frames for each video clip, from which the ATP module selects the feature of \textbf{one} frame to represent the video. Both steps are supervised by KL contrastive loss as Equation~\ref{eq:kl_loss}.

After that, both the frame encoder and ATP module are frozen during the joint training of lite and heavy networks on our One4All-P dataset. For each video in a batch, we sample $L_1\times N$ frames to the frame encoder and the ATP module selects $L_1$ frames. A shallow Transformer will aggregate the frame features to $\mathbf{F}_l$. 
Another $L_2$ frames are sent to the heavy network for video representation $\mathbf{F}_h$. Both features are aligned with the text feature by constraint:
\begin{equation}
\begin{aligned}
\mathcal{L}_{CL}\left(\mathbf{F}_l, \mathbf{F}_h, \mathbf{T}\right) &  = 
    \mathcal{L}_{kl}\left(\mathbf{F}_l , \mathbf{T}, \bm{y}\right) \\
      + \mathcal{L}_{kl} & \left(\mathbf{F}_h  , \mathbf{T}, \bm{y}\right) + \mathcal{L}_{ce}\left(\mathbf{F}_l, \mathbf{F}_h\right).
\end{aligned}
\end{equation} 
$\mathcal{L}_{ce}$ is the CE contrastive loss in CLIP~\cite{clip} since we only align the lite and heavy features of the same instance.

During inference, the lite and heavy networks independently generate prediction by cosine similarity to the text embeddings of the classes. The two streams can be combined by mean pooling to produce the \textit{Full} model result. 
In the non-zero-shot scenario, linear probing can be applied to enhance fixed-class recognition performance.
Thus, our method is flexible in HOI learning. The full model achieves better model performance, while the lite or heavy models are more efficient and amenable to customization.

Given the special properties of Ego-HOI videos, we further design two customized constraints to better utilize the rich information inherent in Ego-HOI videos.

{\bf Serial Visual Scene Attention Learning (SVSA).}
If a model can learn human intention from its associated view changes, dubbed as SVSA, its focus on the view of HOI should be temporally continuous and thus predictable. 
We enhance the learning of SVSA with an auxiliary task by proposing to predict the movement of the view center from the semantic feature flow.
As shown in Figure~\ref{fig:model}, we hope the motion direction can be recoverable.
Thankfully, for each video clip with sampled frame features $\mathbf{F}=\bm{f_i}|_{i=1}^L$, we have already extracted the camera motion $\bm{m}=\left[x, y\right]$ during the dataset analysis, which stands for the movement of the camera center from $L^{th}$ frame to $(L+1)^{th}$ frame.
 So we propose the following SVSA constraint:
\begin{equation}
\begin{aligned}
\mathcal{L}_{SVSA}\left(\mathbf{F}, \bm{m}\right) & = 1-\cos\langle \mathcal{F}_s(\mathbf{F}), \bm{m} \rangle, 
\end{aligned}
\end{equation}
where $\mathcal{F}_s$ is a shallow network receiving a frame feature sequence and outputs a 2D direction vector. The negative cosine drives the predicted angle to approach ground truth. Here, we use 2D motion. Considering 3D may bring a new improvement, but it is more expensive given egocentric videos for 3D reconstruction.

{\bf Counterfactual Reasoning for Ego-HOI.}
Counterfactual causation studies the outcome of an event if the event does not actually occur and we leverage counterfactual learning to enhance causal \textit{robustness}. 
We construct counterfactual Ego-HOI samples. In Figure~\ref{fig:model}, for a clip with frames $\mathbf{F}=\bm{f_i}|_{i=1}^L$, we modify the ``hand'' node (hand state) and expect changes in the output.
We sample $\alpha\%$ from the $L$ frames and construct counterfactual video $\mathbf{F}_{cf}$ by 1) replacing the whole frames by frames in the same video but with dissimilar hand pose or action label, or 2) if possible, replacing the hand area by hand boxes of other frames with different hand poses or action labels. Thus we modify the hand node without changing other nodes.
We propose a constraint to supervise the prediction after counterfactual modification, as a ``reversed'' contrastive loss~\cite{hadsell2006dimensionality}:
\begin{equation}
\mathcal{L}_{CF}(\mathbf{F}, y) = \max\left[0, \gamma - \cos\langle \mathcal{T}(y), \mathcal{V}(\mathbf{F}_{cf}) \rangle \right]^2,
\end{equation}
where $\mathcal{V}$ and $\mathcal{T}$ are the visual (lite/heavy) and text net, and $\gamma$ is the contrastive margin to clamp the cosine similarity and penalize the cosine similarity that is smaller than  $\gamma$. This constraint ensures that the label of the counterfactual sample is semantically different from the original GT.

The full training loss with weight $\lambda_1, \lambda_2$ is:
\begin{equation}
    \mathcal{L} = \mathcal{L}_{CL} + \lambda_1 \mathcal{L}_{SVSA} + \lambda_2 \mathcal{L}_{CF}.
\end{equation}

\subsection{All-for-One (All4One) Customized Mechanism}
\label{sec:method:all4one}

Our pre-train set and baseline yield high-performing Ego-HOI models, which can be further strengthened with customized strategies on each dataset. 
Besides the dataset-specific finetuning, we can add informative samples to enhance the performance with \textbf{minimum overhead} based on the video properties of each instance and video selection (Algorithm~\ref{alg:sampling-algorithom}), instead of adding data optionally.
And recent research~\cite{sener2017active} shows that removing samples only results in minor performance degradation, and at times even produces slight improvement. Thus we apply \textit{data pruning} before \textit{addition} to offset its overhead. The pruning is similar to video selection where instances with high KDE likelihood are removed, as they are more likely to be redundant.

\begin{figure*}[t]

\centering
    \begin{minipage}[b]{0.28\textwidth}
        \centering
        \resizebox{\textwidth}{!}{
            \begin{tabular}{l|c}
                \hline 
                Method                         & Ego4D-AR  \\ 
                \hline
                I3D~\cite{i3d}                 & 14.6*     \\
                SlowFast~\cite{slowfast}       & 16.1*     \\
                ActionCLIP~\cite{actionclip}   & 12.3*     \\
                MViT-B/16x4~\cite{mvit}        & 16.3*      \\
                ViViT-L/16x2~\cite{vivit}      & 16.1*     \\
                \hline
                Ours (lite)                    & 12.7     \\
                Ours (heavy)                   & 17.2     \\
                Ours (full)                    & \textbf{17.6}      \\
                \hline
            \end{tabular}
        } 
        \subcaption{Ego4D-AR}
    \end{minipage}
    \hfill
    \begin{minipage}[b]{0.28\textwidth}
        \centering
        \resizebox{\textwidth}{!}{
            \begin{tabular}{l|c}
                \hline
                Method                         &EPIC-100 \\
                \hline
                TSM~\cite{tsm}                 & 67.9  \\
                Ego-Exo~\cite{ego-exo}         & 67.0 \\
                IPL~\cite{ipl}                 & 68.6 \\
                ViViT-L/16x2~\cite{vivit}      & 66.4 \\
                MFormer-HR~\cite{mformer-hr}   & 67.0 \\
                TimeSformer~\cite{timesformer} & 67.1 \\
                MeMViT/16x4~\cite{memvit}      & \textbf{70.4} \\
                \hline
                Ours (lite)                    & 62.9  \\
                Ours (heavy)                   & 67.9  \\
                Ours (full)                    & 68.7  \\
                \hline
            \end{tabular}
        } 
        \subcaption{EPIC-KITCHENS-100}
    \end{minipage}
    \hfill
    \begin{minipage}[b]{0.24\textwidth}
        \centering
        \resizebox{\textwidth}{!}{
            \begin{tabular}{l|c}
                \hline
                Method                         & EGTEA  \\ 
                \hline
                Kapidis \etal~\cite{kapidis19} & 65.7$^\dagger$  \\
                Min \etal~\cite{min2021}       & 69.6$^\ddagger$  \\
                Zhang \etal~\cite{zhang2022can}& 69.6$^\dagger$  \\
                \hline
                I3D~\cite{i3d}                 & 58.0  \\
                TSM~\cite{tsm}                 & 60.2  \\
                Ego-RNN \etal~\cite{sudha}     & 58.6  \\
                SAP~\cite{SAP}                 & 62.0  \\
                TSM+STAM~\cite{STAM}           & 64.0  \\
                Lu \etal~\cite{lu2019}         & 68.6  \\
                \hline
                Ours (lite)                    &  66.2    \\
                Ours (heavy)                   &  69.8    \\
                Ours (full)                    &  \textbf{70.8} \\
                \hline
            \end{tabular}
        } 
        \subcaption{EGTEA Gaze+ split 3}
    \end{minipage}
    \captionof{table}{Performance comparison on Ego4D-AR, EPIC-KITCHENS-100, EGTEA Gaze+ of the all-for-one models finetuned on each respective dataset. The results with * are reproduced.
    ($\dagger$: accuracy on 3 splits; $\ddagger$: accuracy on split 1. 
    The actual split 3 accuracy is lower than the reported score for these two methods). Top-1 accuracy is reported here.
    } 
    \label{tab:all4one}

\end{figure*}

\section{Experiments}
\label{sec:exp}

\subsection{Datasets}

Our experiments are conducted on several widely-employed egocentric datasets: EPIC-KITCHENS-100~\cite{epic}, EGTEA Gaze+~\cite{egtea}, Ego4D-AR~\cite{ego4d} (Table~\ref{tab:pretrainset}). Please refer to the supplementary for details of Ego4D-AR.
We report top-1 verb accuracy on EPIC-KITCHENS-100, Ego4D-AR, and action accuracy on EGTEA-Gaze+.

\begin{table}[t]
    \centering
    \resizebox{\linewidth}{!}{
        \begin{tabular}{l|cccc}
            \hline
            Method                                 &Ego4D-AR&EGTEA \\
            \hline
            SOTA                                   & 16.3 & 64.0 \\
            Ours (full)                            & \textbf{17.6} & \textbf{70.8} \\
            \hline
            Ours (full, +5\%)                      & 17.8 &  71.1  & \\
            Ours (full, +10\%)                     & \textbf{18.5} &  \textbf{71.5}  \\
            Ours (full, +20\%)                     & 17.9 &  \textbf{71.5}  \\
            \hline
            Ours (full, -5\% $\Rightarrow$ +5\%)   & 17.7  &  \textbf{70.4}  \\
            Ours (full, -10\% $\Rightarrow$ +10\%) & \textbf{17.9}  &  70.2  \\
            Ours (full, -20\% $\Rightarrow$ +20\%) & 16.8  &  68.4  \\
            \hline
        \end{tabular}
    } 
    \captionof{table}{Results of customized all-for-one strategy. Adding samples with our sampling strategy brings significant improvement. Replacing samples also enhances the models while maintaining the training efficiency.}
    \label{tab:all4one_add_remove}
\end{table}

\subsection{Implementation Details}

We apply video selection on several datasets and in particular, Something-Else~\cite{sthelse} is used in pre-train set construction. 
In the analysis and selection, 
for semantics, we use pre-trained BERT-Base for semantic embeddings.
For hand location and pose, the frames are sampled at FPS=2 and we use cascade mask-RCNN with ResNeXt101 and ResNet50 pose estimator~\cite{xiao2018simple} pre-trained on Onehand10k~\cite{wang2018mask} from MMPose~\cite{mmpose}.
For object location, the frames are sampled at FPS=2 and we use ImageNet-21K~\cite{imagenet1}+LVIS~\cite{gupta2019lvis} pre-trained Detic~\cite{detic} with Swin transformer~\cite{swin}. 
For camera motion, the frames are sampled at FPS=8 and we estimate the Gunnar-Farneback optical flow. The shift vectors are put into 90 bins by their angles. 
For blurriness, we resize the frames to 65,536 pixels for a fair comparison.
We use ViT as the lite network and MViT as the heavy network. The frame-level ViT is 12-layered and the patch size used is 16. The video MViT~\cite{mvit} receives $16 \times 16 \times 3$ tubelet embeddings. 
The text network is a 12-layered Transformer.
The ATP module connecting the image stream and video stream is a fully connected layer.
The aggregator of frame features in the lite network is 6-layered Transformers and the SVSA estimator is 3-layered Transformers.

For more details, please refer to the supplementary.

\subsection{One-for-All Model}

We first train our model on One4All-P. We compare our method with multiple methods and pre-train datasets. 
Table~\ref{tab:one4all} shows that our Full method surpasses the previous models and pre-train sets on all benchmarks. The lite network also outperforms previous methods on EGTEA-Gaze+.
The overall performance on Ego4D-AR is lower than other benchmarks due to its zero-shot test samples.

{\bf Pretrain Set Comparison.}
(Experiment \{2, 3, 6\}, \{4, 7\} , and \{8, 10, 11, 12\})
The models trained on One4All-P outperform the counterparts with other pre-train sets such as Kinetics-400. Besides, we randomly sample a subset of Random-50K from the same 4 datasets as One4All-P. Although given the same data source, pretraining on our One4All-P is superior to a random subset, showing that a balanced dataset indeed benefits the pretraining process and the design taken into consideration of our proposed video properties is proper for Ego-HOI videos.

{\bf Model Comparison.}
(Experiment \{6, 7, 9, 12\})
On the same pre-train set One4All-P-50K or Random-50K, our Full is the strongest one-for-all baseline on most benchmarks. On Ego4D-AR, our baseline is comparable to ActionCLIP but outperforms it on other benchmarks. 

{\bf One4All-T.}
Besides the existing benchmarks, we evaluate the models on our test sets One4All-T in different sizes. One4All-T is a more balanced and harder test set, while our Full model still outperforms the rest on our test sets.

\subsection{All-for-One Mechanism}

With the pre-trained model, we finetune the model on each dataset, as shown in Table~\ref{tab:all4one}.
Most baselines adopt Kinetics pretraining, only except TSM and Ego-RNN.
Our method achieves the SOTA on Ego4D-AR, EGTEA-Gaze+ by over 1\% margin. 
On EPIC-100, our method could be further improved if using a larger temporal reception field similar to MeMViT.
We also apply our sampling strategy to select informative samples and strengthen our baseline in Table~\ref{tab:all4one_add_remove}. With our unified selection criterion, adding only 10\% of samples from the data pool brings about a 1\% improvement on all datasets.
Moreover, to enhance our model while keeping the training efficiency, we \textit{replace} part of the train set with video selection and maintain the data size.
Replacing only 5\% to 10\% of samples can bring performance gain on Ego4D-AR without adding cost. While on EGTEA-Gaze+, replacing samples leads to comparable performance, and replacing 20\% leads to a drop since EGTEA has a larger domain gap than the other datasets and it is hard to find substitutes that can compensate for the semantic loss.

\subsection{Ablation Study}

We conduct ablations to justify our modules and designs. 

{\bf SVSA and Counterfactual Reasoning.} We exclude the SVSA or counterfactual reasoning task during training. As shown in Table~\ref{tab:abl}, the full model suffers degradation without either $\mathcal{L}_{SVSA}$ or $\mathcal{L}_{CF}$, which verifies their efficacy based on the unique property of Ego-HOI video.

{\bf Factor Weights.} The factor weights of our unified criterion are crucial in video selection and dataset construction. The weight is derived according to the analysis and experiments on the video properties, where we find hand/object location, hand pose, and camera motion are more important. 
We conduct comparisons of weights in Table~\ref{tab:abl-weight} and we use the empirical best weight combination in our method.

For more visualizations, limitations, and discussions, please refer to our supplementary materials.

\begin{table}[t]
\centering
    \resizebox{0.75\linewidth}{!}{
        \begin{tabular}{l|cc}
            \hline
            Method                   & EGTEA    & Ego4D-AR \\
            \hline
            Full Model               & \textbf{70.8} & \textbf{17.6}\\
            \hline
            w/o $\mathcal{L}_{SVSA}$ & 70.1   &  16.8 \\
            w/o $\mathcal{L}_{CF}$   & 70.3   &  17.0 \\
            w/o linear probing       & 70.6   &  16.8  \\
            \hline
        \end{tabular}
        } 
        \captionof{table}{Ablation verifying model components, weights for video properties for the unified criterion, and model constraints.}
        \label{tab:abl}
\end{table}

\begin{table}[t]
\centering
    \resizebox{0.85\linewidth}{!}{
        \begin{tabular}{l|ccc}
            \hline
            Unified Weight  & +5\% & +10\% & +20\%   \\ 
            \hline
            $1:1:1:1:1:1$   & \underline{70.6} &  70.0 & 70.7    \\
            $0:1:1:0:1:0$   & 70.2 &  70.0 & 70.4    \\
            $0:1:1:1:1:0$   & \textbf{70.8} &  \underline{70.3} & \underline{70.9}    \\
            $5:10:8:8:10:5$ (Ours) & \underline{70.6} &  \textbf{70.6} & \textbf{71.2}    \\
            \hline
        \end{tabular}
        } 
        \captionof{table}{Ablations on weights (semantics, hand box, pose, object box, camera motion, blurriness) of the unified property on EGTEA-Gaze+. Baseline (w/o additional data) accuracy: \textbf{69.4}\%.}
        \label{tab:abl-weight}
\end{table}

\section{Conclusion}
In this work, we propose a new framework for Ego-HOI learning.
Different from previous works relying on tools, data, and mechanisms from the 3rd-view recognition, we provide more balanced pre-train and test sets with more diverse semantics and hand-object spatial configurations to improve the pre-training and evaluation. And we propose a baseline and training mechanisms for downstream tasks. 
The experiments validate that our framework not only achieves SOTA on multiple benchmarks but also paves the way for more robust and fruitful Ego-HOI studies.

{\small
\bibliographystyle{ieee_fullname}
\bibliography{egbib}
}

\clearpage
\twocolumn[\centering\textbf{\Large Supplementary Materials for EgoPCA: A New Framework for Egocentric Hand-Object Interaction Understanding}
\vspace{0.25in}]

\section{Visualization}
\begin{figure}[t]
    \centering
    \includegraphics[width=\linewidth]{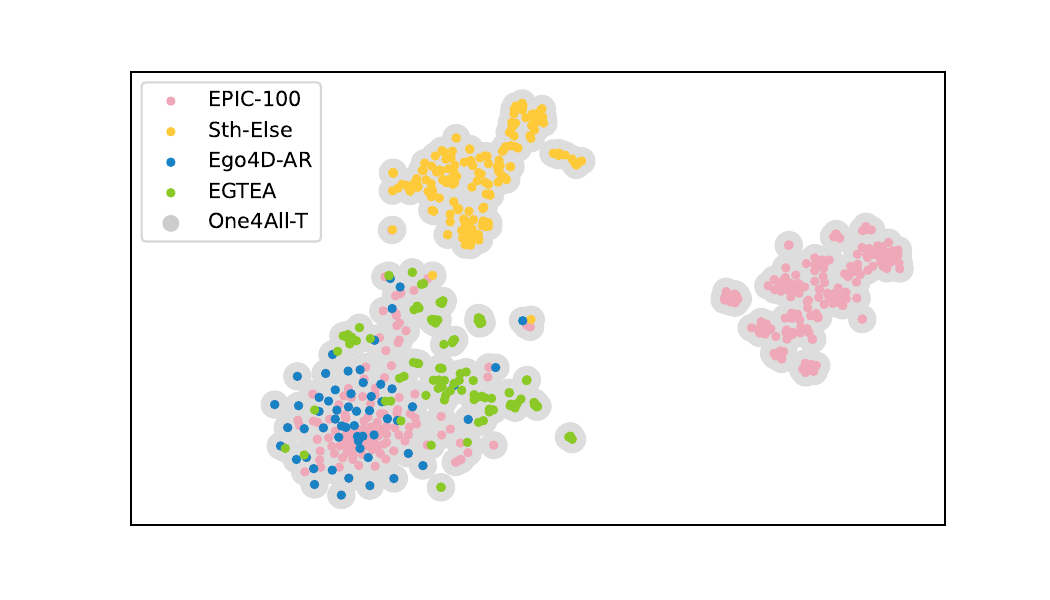}
    \caption{Semantic distribution of actions of Ego-HOI \textbf{test} sets. We use BERT [9] embeddings to visualize the classes.}
    \label{fig:semantic-val}
\end{figure}

\begin{figure}[t]
    \centering
    \includegraphics[width=\linewidth]{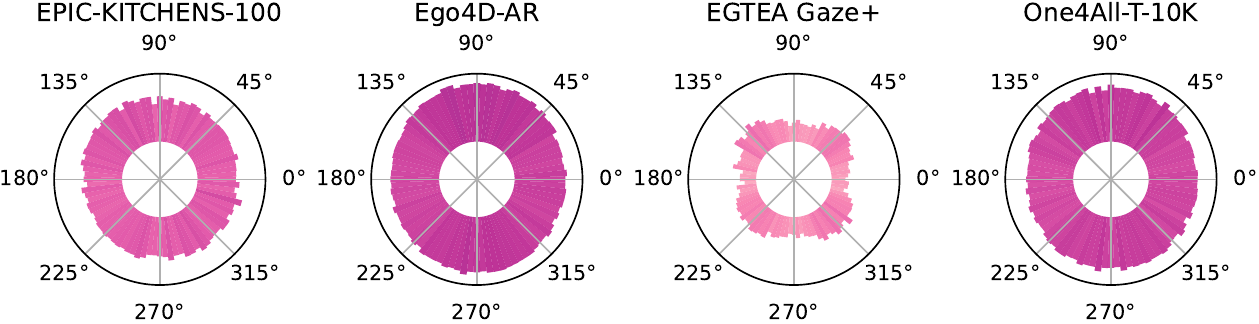}
    \caption{Camera motion polar histogram of Ego-HOI \textbf{test} sets.}
    \label{fig:motion-val}
\end{figure}

\begin{figure}[t]
    \centering
    \includegraphics[width=\linewidth]{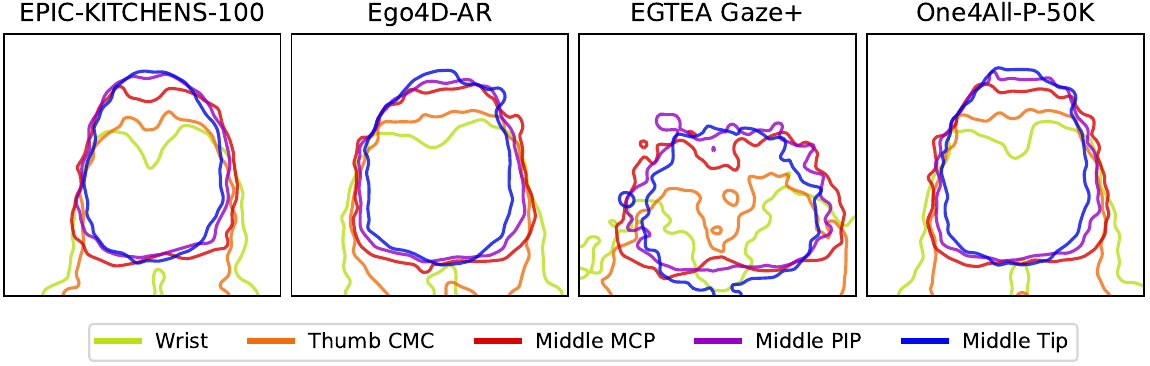}
    \caption{Hand pose. We show the high-density contours of the heatmaps of different hand keypoints on different \textbf{test} sets.}
    \label{fig:hand-pose-val}
\end{figure}

\begin{figure}[t]
    \centering
    \includegraphics[width=\linewidth]{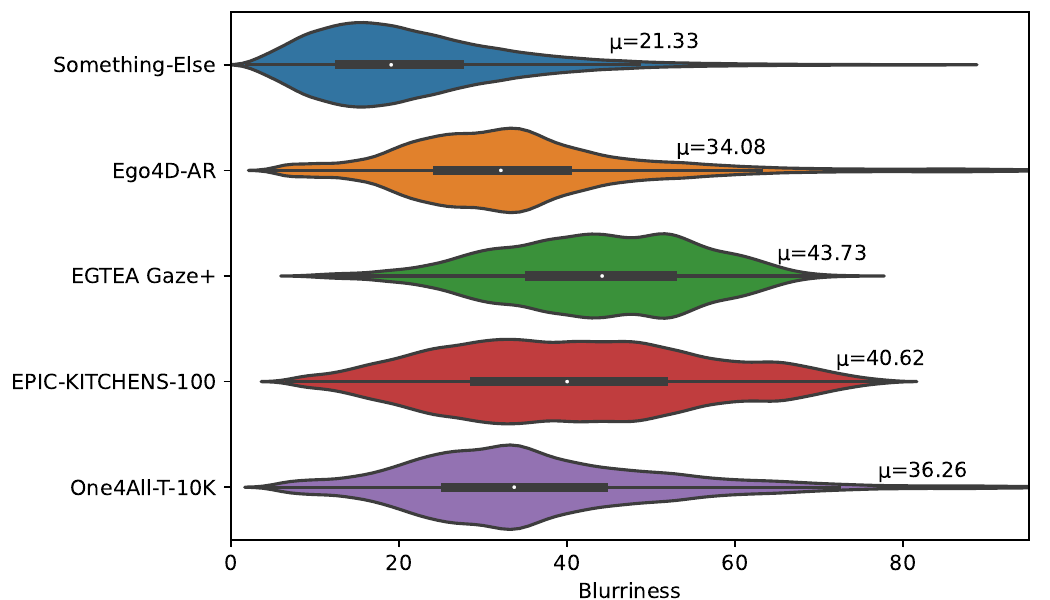}
    \caption{Blurriness (\textbf{test} sets). $\mu$: average blurriness value.}
    \label{fig:blur-val}
\end{figure}

\begin{figure}[t]
    \centering
    \includegraphics[width=\linewidth]{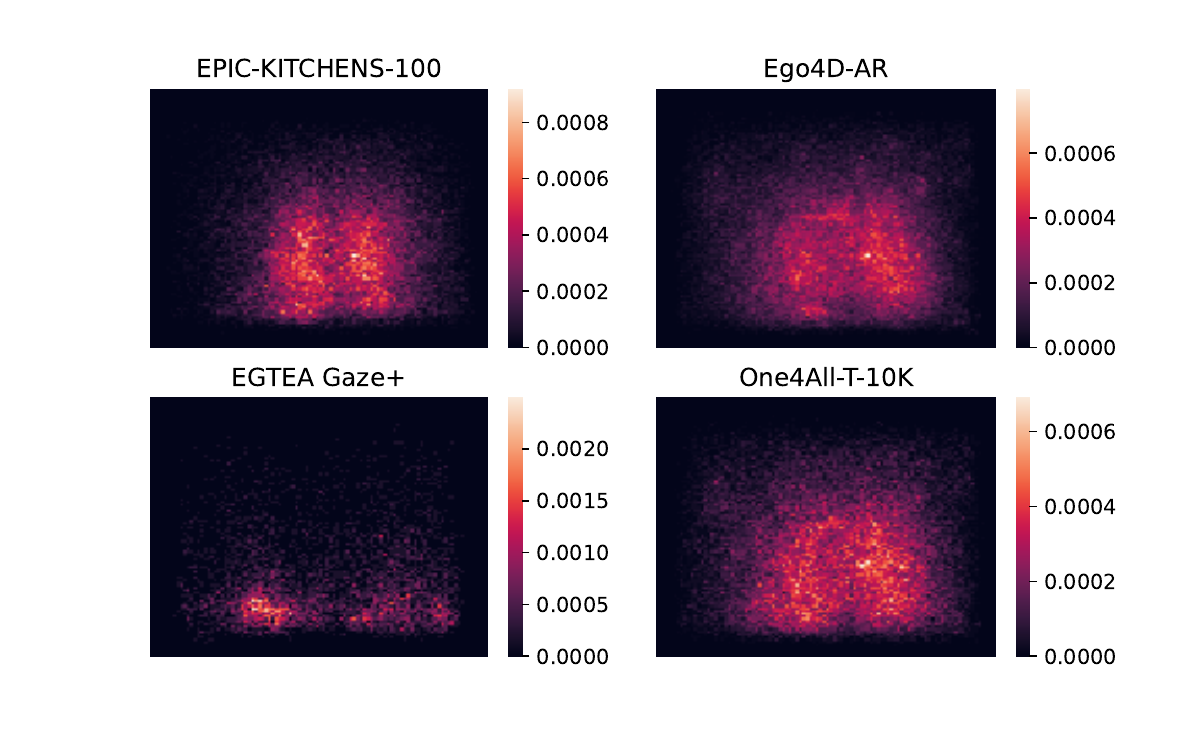}
    \caption{Hand location heatmaps of Ego- HOI datasets (\textbf{test} set).}
    \label{fig:hand-box-val}
\end{figure}

\begin{figure}[t]
    \centering
    \includegraphics[width=\linewidth]{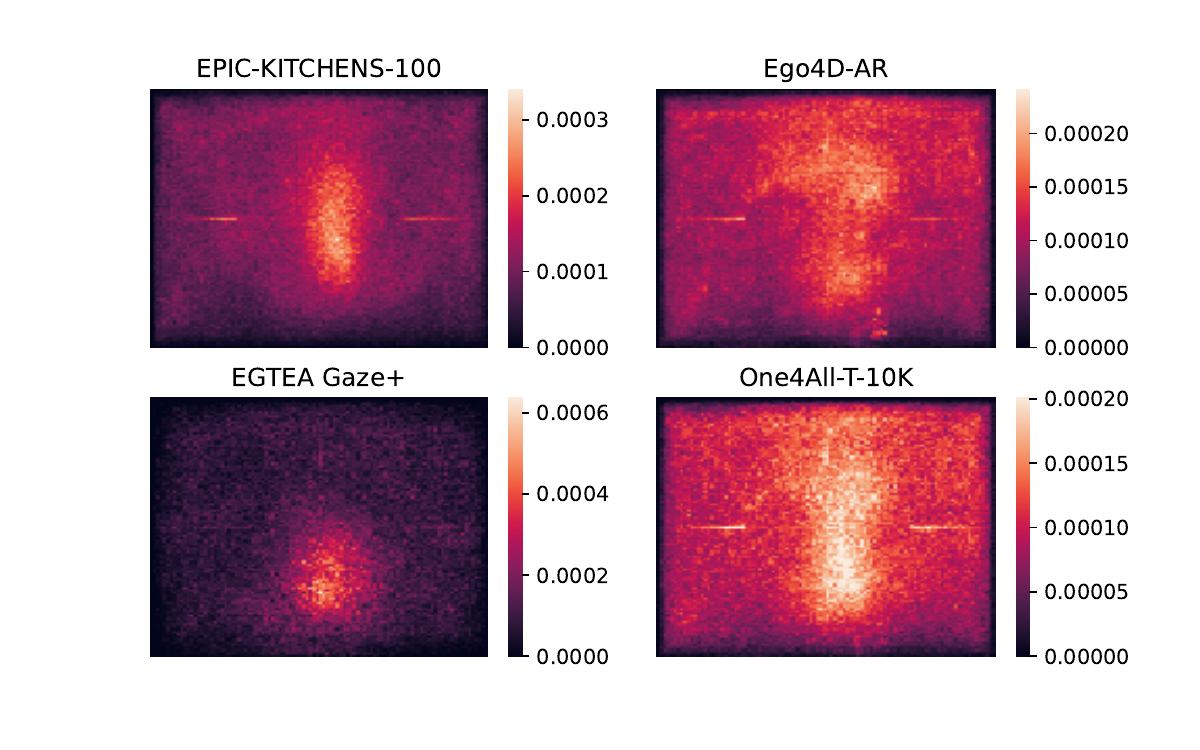}
    \caption{Object location heatmaps of Ego- HOI datasets (\textbf{test} set).}
    \label{fig:obj-box-val}
\end{figure}

\begin{figure}[t]
    \centering
    \includegraphics[width=\linewidth]{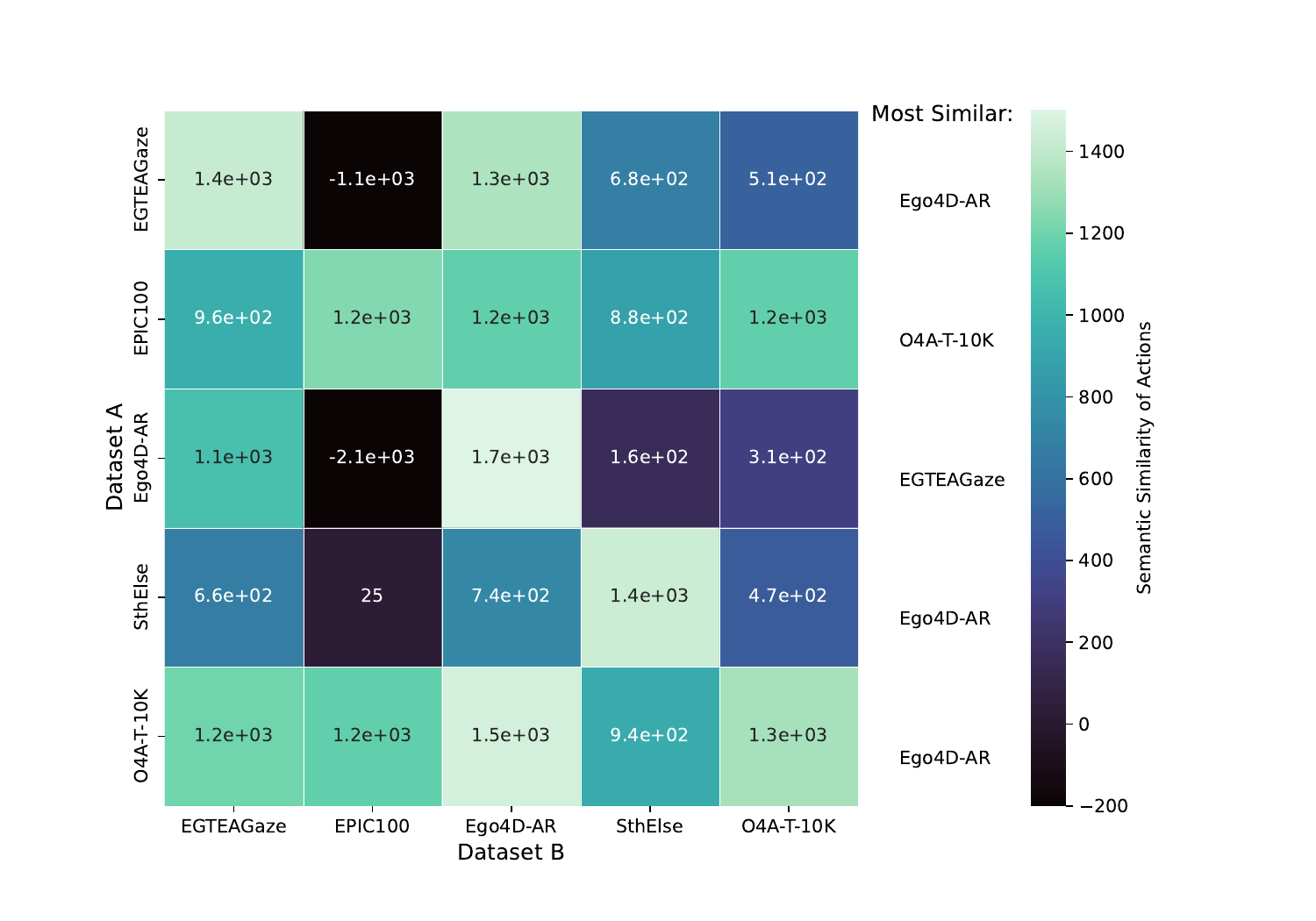}
    \caption{The unified ego-property similarity between \textbf{test} sets.}
    \label{fig:semantic-sim}
\end{figure}

For a more comprehensive analysis, we present more visualizations of video properties:

\noindent{\bf Video Property on The Val/Test Set}.
We present visualizations of the video properties on the test or valid sets, including action semantics (Figure~\ref{fig:semantic-val}), camera motion (Figure~\ref{fig:motion-val}), hand pose (Figure~\ref{fig:hand-pose-val}), blurriness (Figure~\ref{fig:blur-val}), hand box (Figure~\ref{fig:hand-box-val}) and object box (Figure~\ref{fig:obj-box-val}). Our proposed dataset One4All-T is more comprehensive on these properties. We also show the semantic similarity matrix of the test sets in Figure~\ref{fig:semantic-sim}.


\begin{figure}[ht]
    \centering
    \includegraphics[width=\linewidth]{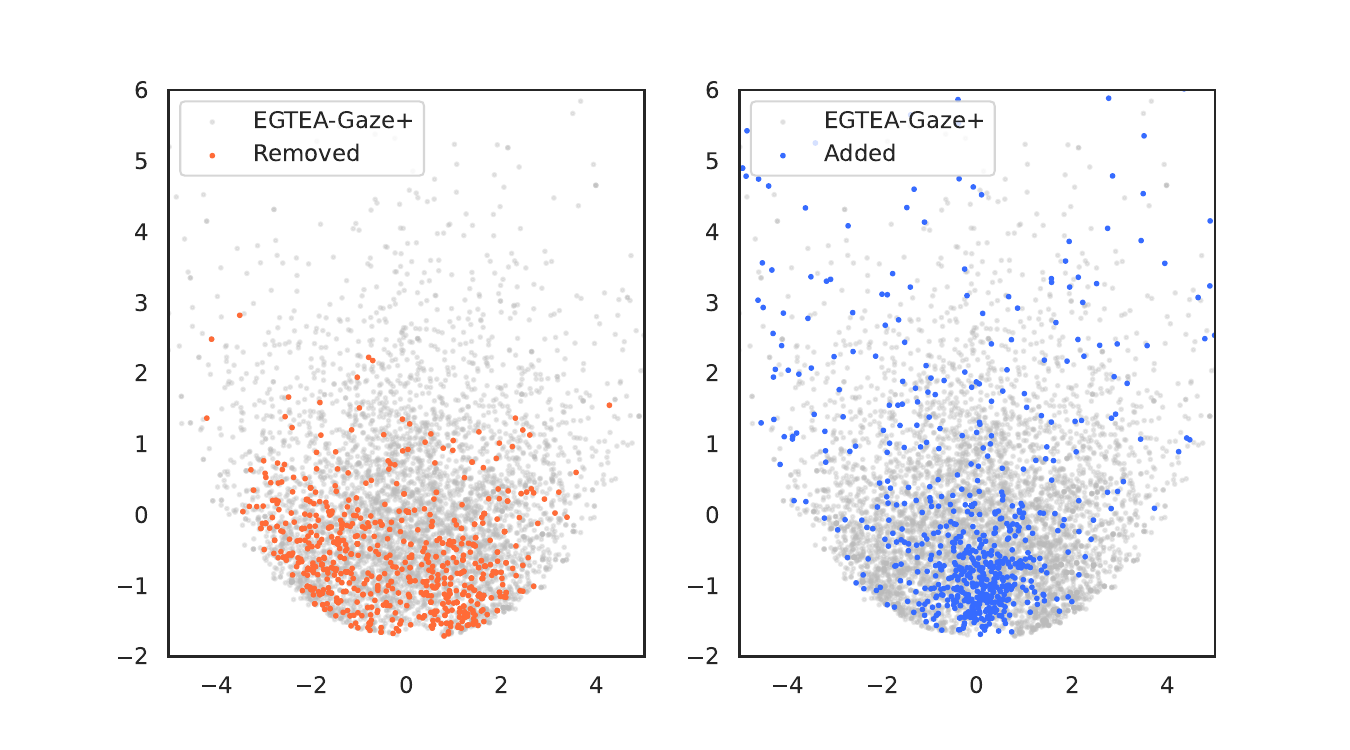}
    \caption{The hand location representations of EGTEA-Gaze+ and the removed (left) or added (right) samples. We use PCA to reduce dimensionality.}
    \label{fig:egtea-handbox}
\end{figure}

\begin{figure}[ht]
    \centering
    \includegraphics[width=\linewidth]{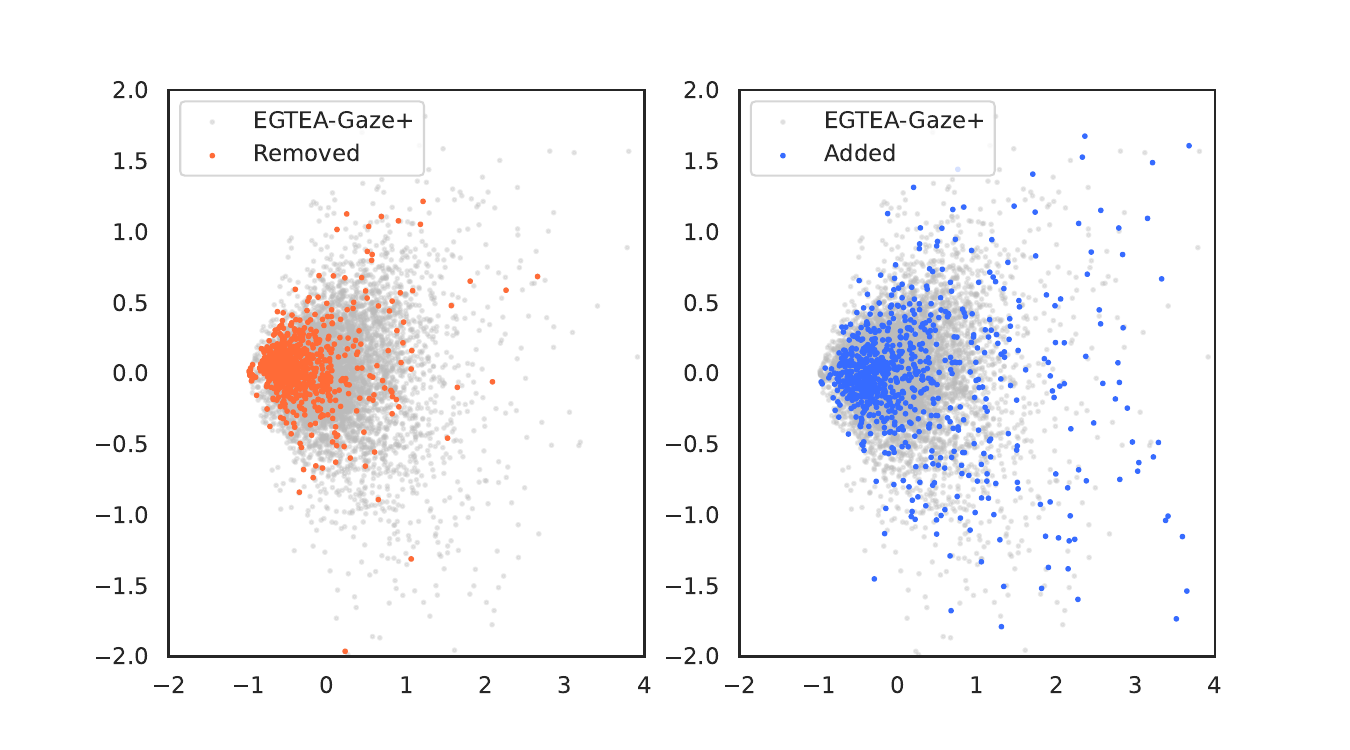}
    \caption{The object location representations of EGTEA-Gaze+ and the removed (left) or added (right) samples. We use PCA to reduce dimensionality.}
    \label{fig:egtea-objbox}
\end{figure}

\paragraph{Data Addition and Removal in All-for-one Setting.}
We visualize the samples added or removed during task-specific model enhancement. 
The data points are represented according to video properties and we show the PCA visualizations of two key properties: hand locations (Figure~\ref{fig:egtea-handbox}) and object locations (\ref{fig:egtea-objbox}).
It can be observed that the removed data usually locate in the dense area of whole datasets. Besides, the additional samples are not only located in the dense area (since our ablation study shows that similar data brings more model improvement) but also supplement more diverse samples located in the sparse regions.


\section{Details of Ego4D-AR}
Ego4D-AR (\textbf{A}ction \textbf{R}ecognition) is derived from the data and annotations of Ego4D~\cite{ego4d}.
We use the annotations from the long-term action anticipation task from Ego4D, which contains start and end frame indices and the corresponding human action. We adopt the video clips from the hand-object interaction task, whose lengths are 8 seconds. A total of 41,085 clips are used for training and 28,348 for validation. The video clips that are exactly within long-term action segments are assigned with action labels, and the rest are discarded.

After the filtering and label assignment, we obtain the \textit{single-label} action recognition dataset Ego4D-AR, which has 22,081 training samples and 14,530 validation samples.
Ego4D-AR has 77 action classes, where 66 are in the train set and 58 are in the validation set. There are 11 \textbf{zero-shot classes} in the validation set due to the filtering process, resulting in the relatively low performance on \textit{One-for-all} task (Table 3 in the main text).

\section{Implementation Details}

\subsection{One4All-P and One4All-T}

We merge the action classes of EPIC KITCHENS 100, EGTEA-Gaze+, Ego4D-AR, and Something-Else and have an active pool with nearly 500 actions. Then the classes with the same semantics are merged. The remaining classes for One4All-P and One4All-T are 394 and 204 respectively (the total number of classes are 401 due to some zero-shot classes).
Then we use our data selection algorithm derived based on the analysis of the video properties presented in the main paper to build our One4All datasets. We respectively sample 20 and 5 instances per class for One4All-P and One4All-T as the initial dataset. Then we select samples according to the unified video property with weight 5:10:8:8:10:5 (\ie, action semantics: hand box: hand pose: object box: camera motion: blurriness). The KDE update frequency $k$ is 2,000/2,500/5,000 for One4All-P-20K/30K/50K and 300/500/1,000 for One4All-T-3K/5K/10K. Note that the larger datasets are built based on the smaller ones (\eg we add 20 K samples to One4All-P-30K to build One4All-P-50K).

\subsection{Model and Training Details}

In the one-for-all training stage, the model is trained with an Adam optimizer with a warmup learning rate of 2.0e-5 for 5 epochs and a cosine learning rate from 1.0e-4 to 1.0e-6 for 90 epochs.
In the all-for-one training stage, the model is finetuned with a learning rate of 5.0e-5.

\subsection{More Ablation Study}

\noindent{\bf Loss weight}
For loss weights $\lambda_1,~\lambda_2$, We select the parameter by cross-validation and the ablation study is given in Table~\ref{tab:abl-lambda}.

\begin{table}[h]
    \centering
    \resizebox{0.7\linewidth}{!}{
        \begin{tabular}{ll|c}
            \hline
            $\lambda_1$  & $\lambda_2$  & EGTEA \\ 
            \hline
            0.2 (default)& 0.1 (default)& \textbf{70.8}  \\
            \hline
            0.5          &     0.1      & \textbf{70.8}  \\
            0.1          &     0.1      & 70.6  \\
            \hline
            0.2          &     0.3      & 70.6  \\
            0.2          &     0.05     & 70.5  \\
            \hline
        \end{tabular}
    } 
    \caption{Ablation study of $\lambda_1$, $\lambda_2$.} 
    \label{tab:abl-lambda}
\end{table}

\noindent{\bf Dataset Components}
Given that Epic-100 and Sth-Else are the main constituents in our pretraining set, we conducted an experiment with pretraining on only these two datasets. The accuracy on EPIC-100/Sth-Else are 54.1\%/44.5\%.

\noindent{\bf Backbone Model on Epic-100}
For a fair comparison to state-of-the-art model on Epic-100, we add an experiment using MeMViT as backbone in our model and achieve 70.7\% accuracy, which is achives SOTA on the benchmark.

\section{Further Discussion}

Currently, we managed to enhance the balancedness of video properties of Ego-HOI datasets with our selection algorithm. But due to the limitation of data diversity and the trade-off between multiple video properties, we can not achieve ideal balancedness on all properties, as shown in Figure~\ref{fig:hand-box-val}, \ref{fig:hand-pose-val}, \ref{fig:obj-box-val}, \ref{fig:motion-val}, and \ref{fig:blur-val}.
In the future, we will extend our video property-based data selection algorithm to \textit{new data collection} and try to use the massive noisy, third-person, or weakly supervised video data.
We will also enhance our baseline and leverage the unique video properties of the Ego-HOI task.

\section{Licences}

The data we use are from the following datasets and are all publicly available and only for research use. Our data pre-processing and selection will be made public.

\begin{itemize}
    \item EPIC KITCHENS 100~\cite{epic}: \hyperlink{https://epic-kitchens.github.io/}{Link}, Creative Commons Attribution-NonCommercial 4.0 International License
    \item EGTEA Gaze+~\cite{egtea}: \hyperlink{https://cbs.ic.gatech.edu/fpv/}{Link}
    \item Ego4D~\cite{ego4d}: \hyperlink{https://ego4d-data.org}{Link}
    \item Something Something~\cite{sthsth}: \hyperlink{https://developer.qualcomm.com/software/ai-datasets/something-something}{Link}
    \item Something Else~\cite{sthelse}: \hyperlink{https://github.com/joaanna/something_else}{Link}
\end{itemize}

And our code is based on the following code repositories. Our code will also be made public.

\begin{itemize}
    \item PySlowFast: \hyperlink{https://github.com/facebookresearch/SlowFast}{Link}, Apache-2.0 License
    \item OpenMMPose: \hyperlink{https://github.com/open-mmlab/mmpose}{Link}, Apache-2.0 License
    \item CLIP: \hyperlink{https://github.com/openai/CLIP}{Link}, MIT license
    \item ActionCLIP: \hyperlink{https://github.com/sallymmx/actionclip}{Link}, MIT License
\end{itemize}

\end{document}